\documentclass[conference]{IEEEtran}
\usepackage{times}
\usepackage[numbers]{natbib}
\usepackage{multicol}
\usepackage{cuted}
\usepackage{caption}
\usepackage[bookmarks=true]{hyperref}
\usepackage{amsfonts}
\usepackage{amsmath}
\usepackage{multirow} 
\usepackage{booktabs}
\usepackage{pgfplots}
\pgfplotsset{compat=1.18}
\usepackage{tcolorbox}
\definecolor{deepPink}{RGB}{255, 20, 147}
\usepackage{tikz}
\usepackage{arydshln}
\usepackage{xcolor}
\renewcommand{\IEEEkeywordsname}{Keywords}
\begin{document}

% paper title
\title{DexGrasp-Zero: A Morphology-Aligned Policy for Zero-Shot Cross-Embodiment Dexterous Grasping}

\author{
  \textbf{Yuliang Wu},
  \, \textbf{Yanhan Lin},
  \, \textbf{WengKit Lao},
  \, \textbf{Yuhao Lin},
  \, \textbf{Yi-Lin Wei},
  \, \textbf{Wei-Shi Zheng},
  \, \textbf{Ancong Wu}\textsuperscript{\,$\dagger$}\\
  School of Computer Science and Engineering, Sun Yat-sen University, China\\
  % \footnotesize{\textsuperscript{$\dagger$} Corresponding author.}\\
  % \href{https://yliangwu.github.io/DexGrasp-Zero/}{\textcolor{deepPink}{https://yliangwu.github.io/DexGrasp-Zero}}
    \footnotesize{\textsuperscript{$\dagger$} Corresponding author.}\\[0.25em]
  {\large\textcolor{deepPink}{\url{https://yliangwu.github.io/DexGrasp-Zero}}}
}

\maketitle
\IEEEpeerreviewmaketitle
% \begin{figure*}[t!]
%     \centering
%     \includegraphics[width=1.0\linewidth,trim=0 11.cm 7cm 0cm,clip]{figs/fig4.pdf}
%     \caption{Real-world experiment examples on three unseen hands.}
%     \label{fig:real_world_experiment}
% \end{figure*}

% \begin{figure*}[t!]
%     \centering
%     \includegraphics[width=1.0\linewidth,trim=0 10.1cm 0cm 0cm,clip]{figs/exp_result.pdf}
%     \caption{\textcolor[HTML]{0080C0}{(a) Simulation results on training hands} and \textcolor[HTML]{0080C0}{(b) real-world deployment on unseen hands} validate the effectiveness and zero-shot transfer capability of our approach.}
%     \label{fig:real_world_experiment}
% \end{figure*}

\begin{strip}
    \centering
   %\vspace{-2em
   \includegraphics[width=\textwidth]{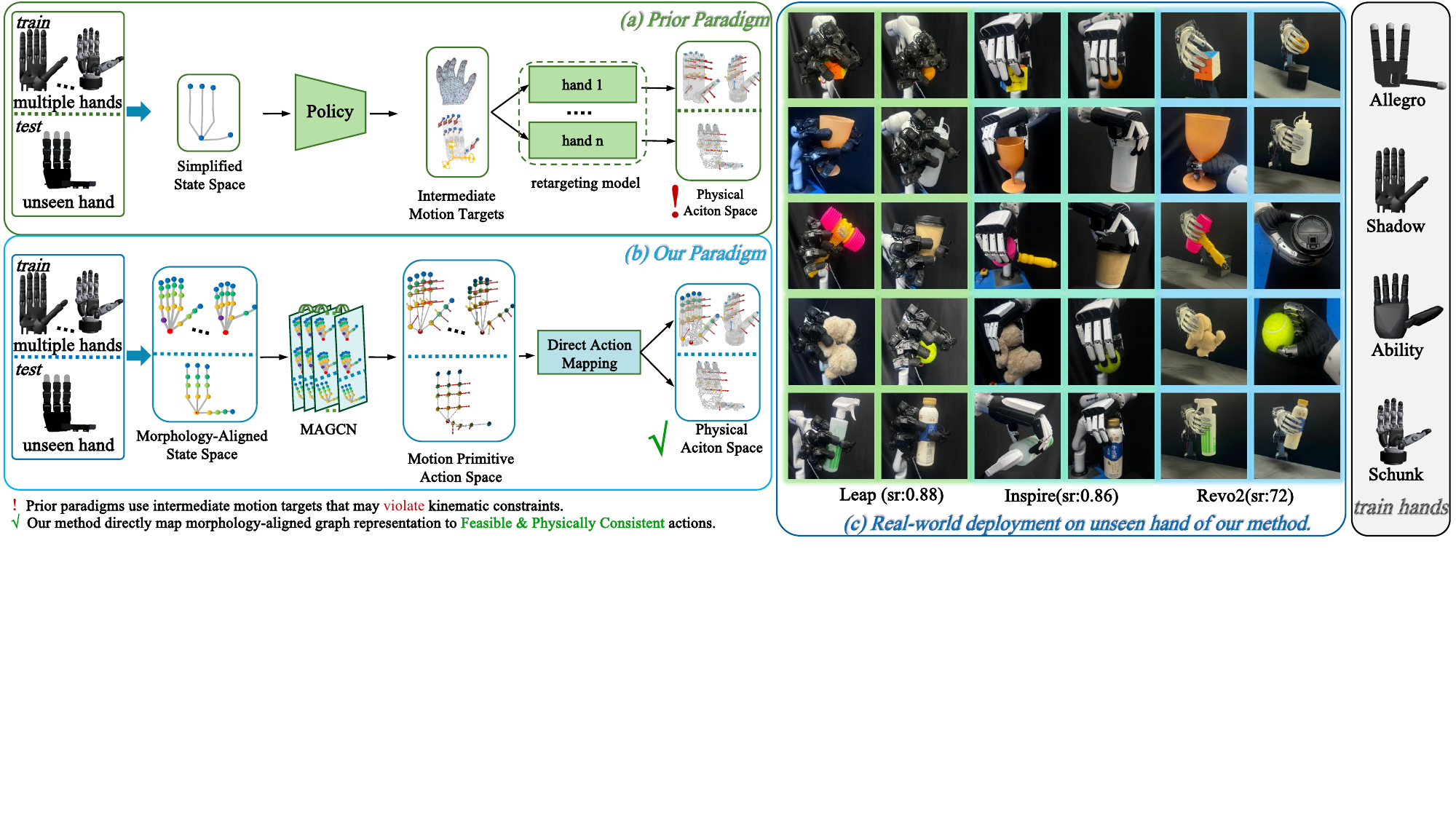}
    \vspace{-3.8cm}
    \captionof{figure}{
    \textbf{Paradigm comparison: prior approaches versus our method.}
    \textcolor[HTML]{4D9134}{(a) Prior paradigm}: Existing methods~\cite{cross,24eccv} train on a simplified and lossy unified state space. They output intermediate motion targets that require hand-specific retargeting models to convert into physical joint commands. This adds complexity and can lead to kinematically infeasible actions.  
    \textcolor[HTML]{00B0F0}{(b) Our paradigm}: We learn a single universal policy end-to-end. The policy operates on a lossless morphology-aligned graph representation and outputs actions in a hand-agnostic motion-primitive space. Physical commands are generated directly through a fixed hand-specific mapping $\mathcal{M}_h$, removing the need for trainable retargeting modules.  
    \textcolor[HTML]{0080C0}{(c) Real-world deployment on unseen hands} validate the effectiveness and zero-shot transfer capability of our approach.
    }
  \label{fig:paradim shift}
\end{strip}

\begin{abstract}
To meet the demands of increasingly diverse dexterous hand hardware, it is crucial to develop a policy that enables zero-shot cross-embodiment grasping without redundant re-learning.
Cross-embodiment alignment is challenging due to heterogeneous hand kinematics and physical constraints. Existing approaches typically predict intermediate motion targets and retarget them to each embodiment, which may introduce errors and violate embodiment-specific limits, hindering transfer across diverse hands.
To overcome these limitations, we propose \textit{DexGrasp-Zero}, a policy that learns universal grasping skills from diverse embodiments, enabling zero-shot transfer to unseen hands.
We first introduce a morphology-aligned graph representation that maps each hand's kinematic keypoints to anatomically grounded nodes and equips each node with tri-axial orthogonal motion primitives, enabling structural and semantic alignment across different morphologies.
Relying on this graph-based representation, we design a \textit{Morphology-Aligned Graph Convolutional Network} (MAGCN) to encode the graph for policy learning. MAGCN incorporates a \textit{Physical Property Injection} mechanism that fuses hand-specific physical constraints into the graph features, enabling adaptive compensation for varying link lengths and actuation limits for precise and stable grasping.
Our extensive simulation evaluations on the YCB dataset demonstrate that our policy, jointly trained on four heterogeneous hands (Allegro, Shadow, Schunk, Ability), achieves an 85\% zero-shot success rate on unseen hardware (LEAP, Inspire), outperforming the state-of-the-art method by 59.5\%. 
Real-world experiments further evaluate our policy on three robot platforms (LEAP, Inspire, Revo2), achieving an 82\% average success rate on unseen objects.

\end{abstract}

\section{Introduction}

In the past several years, industry and academia have rapidly advanced dexterous hand hardware~\cite{zhou2025designadaptivemodularanthropomorphic,weng2025bidexhanddesignevaluationopensource}, and more morphologically heterogeneous hands are expected to emerge for dexterous manipulation. However, existing reinforcement learning (RL) policies are typically limited to the training morphology and fail to generalize across hand types ~\cite{unidexgrasp++,graspxl,wang2025unigrasptransformer,robustdexgrasp,chen2025clutterdexgrasp,FunGrasp,wei2025cyclemanip,lin2025typetele}
, necessitating costly re-training and data collection for every new hand. Therefore, for practical application, we need a transferable policy that enables zero-shot cross-embodiment grasping without redundant re-learning.

The fundamental obstacle in cross-embodiment learning is the morphological differences across hands, which make it difficult to align both the \emph{input} and \emph{output} spaces of a grasping policy. 
The observations and control commands of different dexterous hands are morphology-dependent: observations reflect the number and arrangement of joints and sensors, while control commands are shaped by degrees of freedom and physical properties such as joint limits and link geometry.
This poses significant challenges for mapping the desired action targets across different hands.

Existing cross-embodiment dexterous grasping researches largely follows two routes: (1) open-loop grasp pose generation ~\cite{dro,tro,artigrasp,wei2025omnidexgrasp,wei2024grasp,wei2025afforddexgrasp}, which lacks closed-loop feedback and is less robust to perturbations; and (2) RL-based policy transfer of intermediate action representations~\cite{cross,24eccv} with retargeting. However, as illustrated in Fig.~\ref{fig:paradim shift}, the intermediate motion targets may violate kinematic or actuation constraints of the target hand, limiting zero-shot generalization.

One key observation is that, despite large embodiment variations, joints across different hands share underlying morphological and kinematic regularities. Moreover, the inter-joint relationships can be naturally represented as a graph structure. Motivated by this, we construct a cross-hand transferable policy based on graph neural network in a RL framework that directly outputs joint-level actions. This avoids explicit retargeting and thus overcoming the inherent limitations of previous methods.

To achieve this, we introduce 
%% origin version below:
% \textbf{DexGrasp-Zero}, a universal grasping policy capable of seamless zero-shot transfer to unseen hands.
% To achieve this, we propose the \textbf{Morphology-Aligned Graph Convolutional Network} (MAGCN), which represents any dexterous hand as an anatomically aligned \emph{graph} that matches the hand's kinematic structure, and encode the state space by graph convolution. %yielding a morphology-consistent state space that can be encoded by a graph convolutional network (GCN)~\cite{GCN}.
% To reduce action-space mismatch, we adopt a hand-agnostic motion primitive space grounded in biomechanics~\cite{Santello1998PosturalHS} and map these primitives to each hand's joint commands. To further ensure stable grasping, we extract embodiment physical properties from the Unified Robot Description Format (URDF) and inject them into the graph representations, enabling the policy to adaptively compensate for varying link lengths and actuation limits.
\textbf{DexGrasp-Zero}, a universal grasping policy capable of zero-shot transfer to unseen hands.
Specifically, we represent each hand as an anatomically grounded, morphology-aligned \emph{graph} that matches the hand's kinematic structure, and define a hand-agnostic motion primitive space grounded in biomechanics~\cite{Santello1998PosturalHS} to align control semantics across morphologies.
We then parameterize the policy with the \textbf{Morphology-Aligned Graph Convolutional Network} (MAGCN), which encodes the graph state for policy learning via graph convolution. To further ensure stable grasping, MAGCN incorporates URDF-derived embodiment physical properties and injects them into the graph representations, enabling the policy to adaptively compensate for varying link lengths and actuation limits.

We evaluate DexGrasp-Zero on six diverse dexterous hands. Our experiments show that a single policy, jointly trained on four hands, achieves a \textbf{85\% zero-shot grasping success rate} on unseen hands (Leap, Inspire), outperforming prior methods by \textbf{59.5\%}.
We further evaluate our policy on three heterogeneous physical robots (Leap, Inspire, Revo2), achieving an average success rate of \textbf{82\%} on 10 unseen objects.

\textbf{Our contributions are summarized as follows:}
\begin{enumerate}
    \item We propose a morphology-aligned graph state representation and a hand-agnostic motion primitive space that align perception and control semantics across heterogeneous dexterous hands.
    
    % \item We design a GCN-based representation learning policy and inject URDF-derived physical properties into the learned graph features to better respect embodiment constraints.
    \item We design MAGCN, a GCN-based policy that injects URDF-derived physical properties into the learned graph features to better respect embodiment constraints.

    \item We conduct extensive experiments in simulation and on three real-robot hands (Leap, Inspire, Revo2), validating the efficacy of our representation and framework for zero-shot cross-embodiment grasping on novel objects.
\end{enumerate}
 We will release code and datasets to support the community’s pursuit of general-purpose robotic manipulation.
 % This work advances cross-embodiment dexterous grasping from engineering heuristics toward a scientific paradigm.
\section{Related work}
\subsection{Learning-Based Dexterous Grasping}
% Dexterous grasping constitutes a foundational capability for robotic interaction with the physical world. Many  studies~\cite{jiang2021hand,unidexgrasp,lu2024ugg,zhang2024dexgraspnet,xu2024dexterous,zhong2025dexgrasp} employ end-to-end generative frameworks to model high-dimensional grasp pose distributions, enabling diverse grasps within a single hand morphology or use reinforcement learning methods ~\cite{unidexgrasp++,wang2025unigrasptransformer,pavlichenko2025dexterous,graspxl,robustdexgrasp,chen2025clutterdexgrasp} to optimize action decisions through environmental feedback in closed-loop interactions, achieving robust grasping. However, the methods relying on generative models are dependent on the grasping data used during training, if different hand types are not covered, cross-hand generalization cannot be achieved. Similarly, reinforcement learning methods, if the state space design and network architecture do not consider different hand types, the trained policies will be bound to specific hand types and also fail to achieve cross-hand generalization.

Dexterous grasping is a cornerstone of robotic manipulation. Prior work employs either generative models~\cite{jiang2021hand,unidexgrasp,lu2024ugg,zhang2024dexgraspnet,xu2024dexterous,zhong2025dexgrasp} to sample diverse grasps or reinforcement learning (RL)~\cite{unidexgrasp++,wang2025unigrasptransformer,pavlichenko2025dexterous,graspxl,robustdexgrasp,chen2025clutterdexgrasp} for closed-loop policy optimization. However, both paradigms typically assume a fixed hand morphology: generative approaches are limited by the hand coverage in training data, and RL policies are often constrained by hand-specific state representations and network architectures, so neither generalizes across hand embodiments without retraining.

\subsection{Cross-Embodied Dexterous Grasping}
To enable cross-hand transfer under morphological heterogeneity, recent efforts often introduce intermediate representations that are less sensitive to hand-specific degrees of freedom and physical properties. Existing methods largely follow two routes: generating static grasping postures, or learning dynamic closed-loop grasping policies.
Existing grasp synthesis methods can be broadly categorized as object-centric~\cite{unigrasp,attarian2023geometry,li2022gendexgrasp,xu2024manifoundation,fang2025anydexgrasp} or obj-hand interaction~\cite{dro,tro}, most of which generate static, kinematically feasible postures and struggle with real-time perturbations. 

Another approach is to train closed-loop grasping strategies that execute complete grasping trajectories. A common design is to use a simplified intermediate action target (e.g., fingertip displacements~\cite{24eccv} or MANO poses~\cite{cross}) and then retarget it to each hand. The subsequent hand-specific retargeting step can produce infeasible joint commands under kinematic constraints. In contrast, our approach enables end-to-end policy learning that maps observations directly to physical joint commands, thereby avoiding kinematic infeasibility.

\subsection{Graph Representation for Cross-Embodied Grasping}
Graph neural networks provide a natural abstraction for dexterous hands, but prior graph-based approaches can suffer from misalignment in cross-hand representations, resulting in a lossy unified interface. GeoMatch++~\cite{geomatch++} uses kinematic links as graph nodes, which lacks anatomical semantics and does not explicitly encode topological connectivity. CrossDex~\cite{cross} and She et al.~\cite{24eccv} model only sparse keypoints without encoding topological relationships, leading to information loss. GET-Zero~\cite{getzero} builds a full kinematic graph but treats each physical joint as a node, splitting a single anatomical unit with multiple dofs into multiple nodes, which hinders semantic alignment across hands with different DoF distributions, and its evaluations are therefore limited to LEAP Hand variants.

In contrast, our method adopts an anatomically aligned and topology-aware graph representation that preserves task-relevant structure across heterogeneous hands. Together with hand-agnostic motion primitives, it aligns both state and action semantics across embodiments, enabling more direct and transferable policy learning.

\section{DexGrasp-Zero}
\label{sec:method}
\begin{figure*}[ht!]
    \centering
    \includegraphics[width=1 \linewidth,trim=2.cm 4cm 1.5cm 0cm,clip]{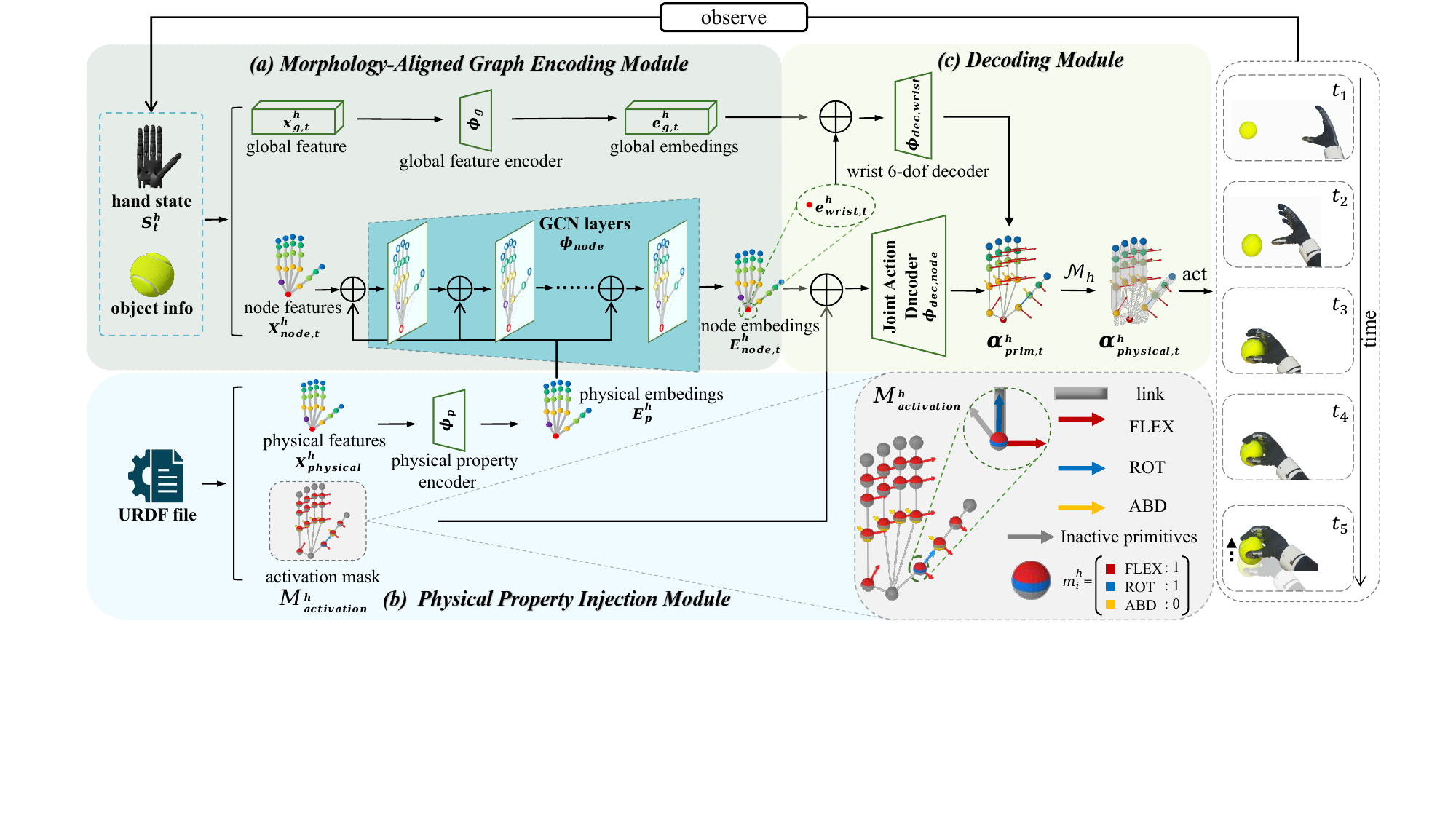}
    \caption{\textbf{Architecture of DexGrasp-Zero.} At each time step \( t \):  
(a) \textbf{Morphology-Aligned Graph Encoder} encodes hand-object state into node features \( \mathbf{X}^{h}_{\text{node},t} \) and global feature \( \mathbf{x}_{g,t}^{h} \) using a hand-specific graph (adjacency \( \mathbf{A}^h \)); a GCN with per-layer physical priors produces embeddings \( \mathbf{E}_{\text{node},t}^{h} \) and \( \mathbf{E}_{g,t}^{h} \).  
(b) \textbf{Physical Property Encoder} parses hand URDF to build a physical graph \( \mathcal{G}_{\text{physical}}^{h} \) (joint limits, link lengths, etc.) and an activation mask \( \mathbf{M}_{\text{activation}}^{h} \), encoded into \( \mathbf{E}_{\text{p}}^{h} \) and fused into every GCN layer.  
(c) \textbf{Decoder} outputs motion primitives \( \boldsymbol{\alpha}_{\text{prim}}^{h} \): wrist 6-DoF commands from \( \mathbf{E}_{g,t}^{h} \) and wrist features, and joint actions from masked node embeddings; the latter are mapped via hand-specific \( \mathcal{M}_h \) to executable joint commands \( \alpha_{\text{physical},t}^{h} \).}
    \vspace{-1em}
    \label{fig:overview}
\end{figure*}

We propose \textbf{DexGrasp-Zero}, a morphology-aligned graph policy that transfers \emph{one} grasping controller across diverse dexterous hands by explicitly extracting \emph{shared} semantics from \emph{hand-specific} mechanics. As shown in Fig.~\ref{fig:overview}, we represent each hand with a morphology-aligned state graph, encode embodiment-specific physical priors from URDF, and fuse them via a morphology-aligned GCN to output a hand-agnostic motion-primitive space that is deterministically mapped to executable joint commands.

\begin{figure}
    \centering
    \vspace{-0.5em}
    \includegraphics[width=0.9\linewidth,trim=0 5cm 14cm 0cm,clip]{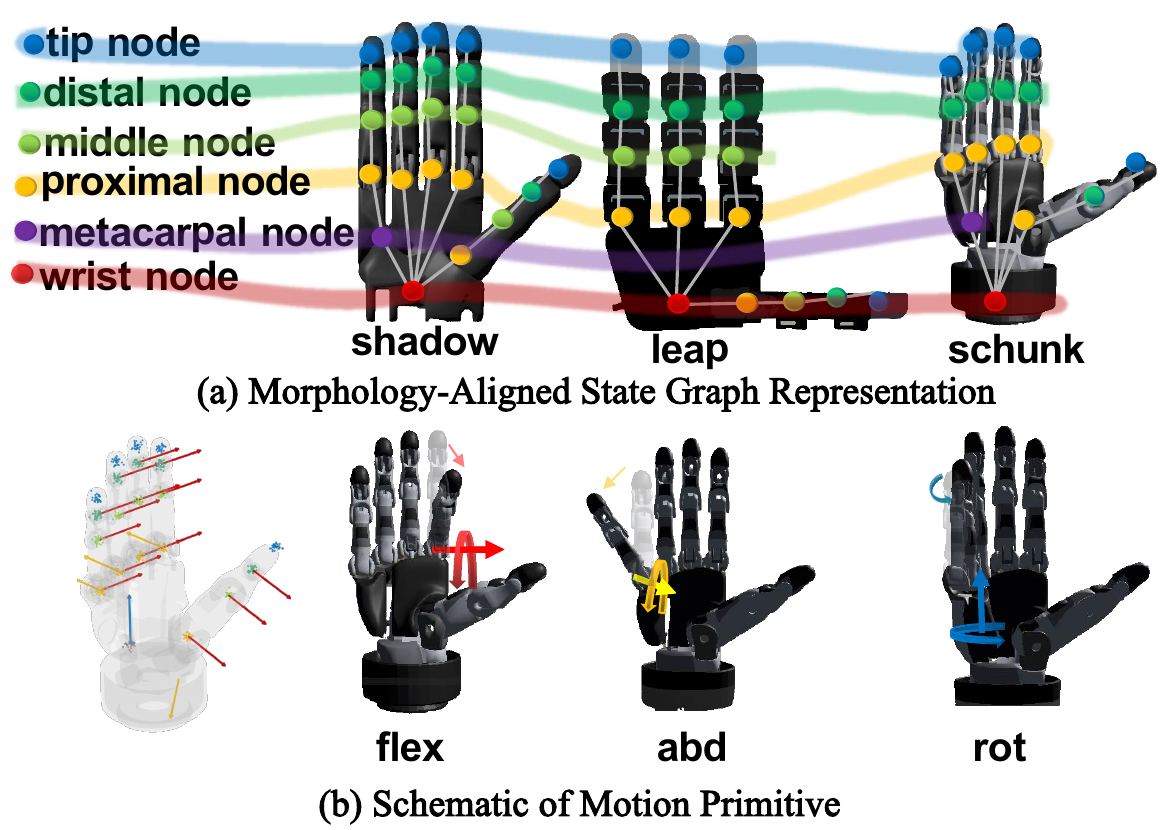}
    \vspace{-0.5em}
    \caption{Universal hand representation. (a) Morphology-Aligned State Graph Representation: nodes correspond to anatomical units, edges follow kinematic chains, yielding a hand-agnostic semantic graph structure.  
(b) Schematic of three motion primitives (Flexion, Abduction, Axial Rotation) on a Schunk hand, showing their physical motion effects at representative joints.}
    \vspace{-0.5em}
    \label{fig:graph_representation}
\end{figure}

\subsection{Problem Formulation for Zero-Shot Cross-Embodiment Grasping}

We consider a unified cross-embodiment learning setting. Let $\mathcal{H}_{\text{train}}$ denote the set of training hands (embodiments) and $\mathcal{H}_{\text{test}}$ denote a disjoint set of unseen test hands. For each hand $h\in\mathcal{H}_{\text{train}}\cup\mathcal{H}_{\text{test}}$, we define a hand-conditioned MDP in which the agent observes a hand-specific state $\mathbf{s}_t^h$ and outputs an action at every time step $t$. Our goal is to learn a \emph{single} shared policy $\pi_\theta$ on $\mathcal{H}_{\text{train}}$ and evaluate it \emph{zero-shot} on $\mathcal{H}_{\text{test}}$ without any finetuning.

Crucially, $\pi_\theta$ outputs an intermediate action representation in a hand-agnostic action space, while a fixed, hand-specific mapping $\mathcal{M}_h$ deterministically converts it into an executable physical command for embodiment $h$. The cross-embodiment control loop is:
\begin{equation}
\boldsymbol{\alpha}_t^h \sim \pi_\theta\big(\cdot\mid \mathbf{s}_t^h\big),
\qquad
\boldsymbol{\alpha}_{\text{physical},t}^h = \mathcal{M}_h\!\left(\boldsymbol{\alpha}_t^h\right),
\label{eq:zero_control_loop}
\end{equation}
where $\boldsymbol{\alpha}_t^h$  denotes the hand-agnostic intermediate action and $\boldsymbol{\alpha}_{\text{physical},t}^h$ is the physical command executed on hand $h$. The learning objective is to maximize the expected discounted return across training hands:
\begin{equation}
    \max_\theta \; \mathbb{E}_{h \sim \mathcal{H}_{\text{train}}} \left[ \sum_{t=0}^{T} \gamma^t\, r\!\left(\mathbf{s}_t^h,\, \boldsymbol{\alpha}_{\text{physical},t}^{h}\right) \right],
\end{equation}
with the requirement that the resulting policy generalizes zero-shot to unseen hands $h'\in\mathcal{H}_{\text{test}}$ at test time.

In the remainder of this section, we instantiate $\mathbf{s}_t^h$ and $\boldsymbol{\alpha}_t^h$ using a morphology-aligned graph representation and a universal motion-primitive space, and specify how $\mathcal{M}_h$ is constructed from the hand kinematics.

\subsection{Morphology-Aligned State and Action Graph Representation}
Prior cross-embodiment grasping systems often struggle with representation mismatch at both the state and action levels: state encodings are often not semantically aligned across morphologies, and actions are frequently output in a task-space 3D targets and require retargeting, which can introduce execution error and yield physically infeasible targets. These issues motivate us to design an morphology-aligned, information-preserving state space and a physically constrained action space that can be executed directly on each hand, avoiding retargeting.

\subsubsection{Morphology-Aligned State Representation}
Dexterous hands exhibit substantial variation in kinematic topology and DoF counts, hindering direct cross-embodiment policy transfer. Our key insight is that, despite these differences, hand kinematics share a set of anatomically meaningful functional units~\cite{Neumann2009KinesiologyOT}. We therefore abstract any dexterous hand $h$ into a morphology-aligned state graph $\mathcal{G}_h = (\mathcal{V}_h, \mathcal{E}_h)$, where $|\mathcal{V}_h|=N_h$ may vary across embodiments and each node falls into one of six semantic types: fingertip, distal, middle, proximal, metacarpal, or wrist. Edges $\mathcal{E}_h$ encode the kinematic relationships between these units, as shown in Fig.~\ref{fig:graph_representation}.

For each node $v_i$, we define a feature vector $\mathbf{x}_i^h \in \mathbb{R}^{d_{\text{node}}}$ capturing its dynamic state. Stacking all node features yields the \textbf{node-feature matrix}
\begin{equation}
\mathbf{X}_{\text{node}}^{h} = [\mathbf{x}_1^{h},\dots,\mathbf{x}_{N_h}^{h}]^\top \in \mathbb{R}^{N_h\times d_{\text{node}}}.
\end{equation}
The hand's kinematic structure is encoded in an adjacency matrix $ \mathbf{A}^h \in \{0,1\}^{N_h \times N_h} $ , where $ \mathbf{A}^h_{ij}=1 $ if node $ i $ is the parent of node $ j $ in the kinematic chain. 

While node features encode local articulation and contact, they lack explicit information about global hand--object relationships critical for coordinated grasping, such as wrist pose error relative to the target or object motion. To address this, we introduce a \textbf{global feature} $ \mathbf{x}_g^h \in \mathbb{R}^{d_{\text{global}}} $ that provides task-level context.  The full state representation is then:
\begin{equation}
    \mathbf{s}^h = \left( \mathbf{X}_{\text{node}}^{h},\, \mathbf{A}^h,\, \mathbf{x}_g^h \right).
\end{equation}

\subsubsection{Hand-Agnostic Motion-Primitive Space}
\label{subsubsec:primitive_action_space}
To achieve hand-agnostic control, we define a universal motion-primitive space $\boldsymbol{\alpha}_{\text{prim}}^h$ based on the morphology-aligned graph to align the control-semantic for each hand. It consists of two parts: (1) a 6-DoF wrist motion command, parameterized as $ (\Delta \mathbf{p}_w^h, \Delta \boldsymbol{\theta}_w^h) $ .  Here and throughout the paper, the symbol $ \Delta $ denotes an incremental change . Specifically, $ \mathbf{p}_w \in \mathbb{R}^3 $ represents the Cartesian position of the wrist, and $ \boldsymbol{\theta}_w \in \mathbb{R}^3 $ represents its orientation expressed in Euler angles. Thus, $ \Delta \mathbf{p}_w^h $ and $ \Delta \boldsymbol{\theta}_w^h $ are the desired translational and rotational displacements of the wrist, respectively.(2) node-level articulation commands defined over the hand-specific state graph $ \mathcal{G}_h = (\mathcal{V}_h, \mathcal{E}_h) $ .
For each node $ v_i \in \mathcal{V}_h $ , we define three orthogonal motion primitives inspired by human hand biomechanics~\cite{Santello1998PosturalHS}, as illustrated in Fig.~\ref{fig:graph_representation}b:
\begin{itemize}
    \item \textbf{Flexion (FLEX)}: rotation that drives the child link toward the normal direction of the palm (i.e., bending inward toward the palm).
    \item \textbf{Abduction (ABD)}: in-plane spreading motion of the finger within the hand plane, moving the digit away from the middle finger axis;
    \item \textbf{Axial Rotation (ROT)}: torsional rotation of the link about its own longitudinal axis.
\end{itemize}

These primitives form a motion-primitive vector for node $v_i$:
\begin{equation}
\boldsymbol{\alpha}^h_i = [\Delta_{\text{flex}},\, \Delta_{\text{abd}},\, \Delta_{\text{rot}}]^\top \in \mathbb{R}^3.
\end{equation}

For notational convenience, we denote the component of $ \boldsymbol{\alpha}_i^h $ corresponding to primitive $p\in\{\text{FLEX},\text{ABD},\text{ROT}\}$ as $\alpha_{i,(p)}^{h}$. 
Stacking wrist 6-DoF commands and all nodes yields the full motion-primitive space $\boldsymbol{\alpha}_{\text{prim}}^h$ for hand $h$:
\begin{equation}
\boldsymbol{\alpha}_{\text{prim}}^h = [\Delta {\mathbf{p}_w^h}^\top,\, \Delta {\boldsymbol{\theta}_w^h}^\top,\, {\boldsymbol{\alpha}_1^h}^\top,\,\dots,\,{\boldsymbol{\alpha}_{N_h}^h}^\top]^\top \in \mathbb{R}^{6+3N_h}.
\label{eq:alpha_prim_stack}
\end{equation}
This universal motion-primitive space achieves control-semantic alignment across heterogeneous hands.

\subsubsection{Mapping to Executable Physical Commands}
\label{subsubsec:mapping_physical_commands}
% To execute the hand-agnostic motion-primitive space on a specific embodiment, we define a fixed, hand-specific mapping $ \mathcal{M}_h $ that projects $ \boldsymbol{\alpha}_{\text{prim}}^h $ onto the embodiment-specific physical joint command space. Specifically, $ \mathcal{M}_h: \mathbb{R}^{6 + 3N_h} \to \mathbb{R}^{L_h} $ is a deterministic, sparse linear operator constructed from the hand's kinematic specification, which:
% \begin{itemize}
%     \item selects valid node--primitive pairs for each physical DoF,
%     \item respects the ordering of joints in the robot's command interface,
%     \item applies a sign correction to align the motion-primitive direction with the physical joint axis.
% \end{itemize}
To execute the hand-agnostic motion-primitive space on a specific embodiment, we define a fixed, hand-specific mapping $ \mathcal{M}_h $ that projects $ \boldsymbol{\alpha}_{\text{prim}}^h $ onto the embodiment-specific physical joint command space. Specifically, $ \mathcal{M}_h: \mathbb{R}^{6 + 3N_h} \to \mathbb{R}^{L_h} $ is a deterministic, sparse linear operator constructed from the hand's kinematic specification, which (1) selects valid node--primitive pairs for each physical DoF, (2) respects the ordering of joints in the robot's command interface, and (3) applies a sign correction to align the motion-primitive direction with the physical joint axis. The physical joint displacement vector is then obtained as
\begin{equation}
    \Delta \mathbf{q}^h = \mathcal{M}_h \big( \boldsymbol{\alpha}_{\text{prim}}^h \big).
\end{equation}

In practice, $ \mathcal{M}_h$ is implemented as an indexing rule; for the $j$-th joint ($ j=1,\dots,L_h$), it satisfies
\begin{equation}
    \Delta q^h_j = s_j^h \cdot \alpha_{n_j,(p_j)}^{h},
\end{equation}
where $n_j\in\{1,\dots,N_h\}$ is the source node, $p_j$ is the selected primitive, and $s_j\in\{-1,+1\}$ is the alignment sign.

We obtain $\mathcal{M}_h$ by applying unit joint excitations in simulation and matching the induced motion semantics to the corresponding motion primitives, which is easy to implement as an indexing rule (see Supplementary Sec.~\ref{subsec:supp_mapping} and Fig.~\ref{fig:primitive_construct}).

The complete physical action executed by the controller combines wrist motion and joint commands:
\begin{equation}
    \boldsymbol{\alpha}^h_{\text{physical}} = \left[ \Delta \mathbf{p}_w^\top,\, \Delta \boldsymbol{\theta}_w^\top,\, \Delta {\mathbf{q}^{h}}^{\top} \right]^\top.
\end{equation}

This hand-specific mapping bridges the universal motion-primitive space and the embodiment-specific command interface, enabling a single shared policy to transfer across hands by acting in a consistent action space.

\subsection{DexGrasp-Zero Policy Design for Dexterous Grasping}
\label{subsec:dex_policy_learning}
Building on the above morphology-aligned state/action abstractions, we design the DexGrasp-Zero grasping policy by specifying (i) a grasping state graph (Fig.~\ref{fig:overview}a), (ii) a URDF-derived physical-property graph (Fig.~\ref{fig:overview}b), (iii) MAGCN with layer-wise physical fusion (Fig.~\ref{fig:overview}a,c), and (iv) the reward.

\subsubsection{Graph-based State Representation}
\label{subsubsec:Graph-based State Representation}
To represent the state in a graph-compatible format, we construct a \textit{state-graph topology} $\mathcal{G}_{h} = (\mathcal{V}_{h}, \mathcal{E}_{h})$ for grasping observations. Our observation design follows RobustDexGrasp~\cite{robustdexgrasp}. The time-varying observation is carried by node/global features (i.e., $\mathbf{X}^{h}_{\text{node}}$ and $\mathbf{x}_g^{h}$), while the topology is encoded by the adjacency $\mathbf{A}^h$.
The state graph  is represented by three components: node features, edge connectivity, and global features.

\textbf{Node Features Matrix ($\mathbf{X}^{h}_{\text{node}}$)}: Given a hand $h$, each node $v_i \in \mathcal{V}_{h}$ is characterized by a feature vector $\mathbf{x}_i^{h} \in \mathbb{R}^{ d_{\text{node}}}$ as:
\begin{equation}
\mathbf{x}_i^{h} = [\mathbf{d}_i^{h},\, \boldsymbol{\theta}_i^{h},\, \dot{\boldsymbol{\theta}}_i^{h},\, c_i^{h},\, f_i^{h},\, \mathbf{m}_i^{h},\, \mathbf{n}_i^{h}],
\end{equation}
where $\mathbf{d}_i^{h} \in \mathbb{R}^3$ is the distance vector from the node to the closest point on the object surface, which is computed from the observed object point cloud by nearest-neighbor search.
$\boldsymbol{\theta}_i^{h},\, \dot{\boldsymbol{\theta}}_i^{h} \in \mathbb{R}^3$ are joint angles and velocities expressed in each the motion-primitive axis; $c_i^{h} \in \{0,1\}$ indicates contact and $f_i^{h} \in \mathbb{R}_{\ge 0}$ is the contact force magnitude. $\mathbf{m}_i^{h}$, $\mathbf{n}_i^{h}$ are one-hot encodings of the finger semantic class (thumb, index, middle, ring, little, wrist) and node type (fingertip, distal, middle, proximal, metacarpal, wrist) respectively. Stacking all node features yields the node-feature matrix $\mathbf{X}^{h}_{\text{node}}=[\mathbf{x}_1^{h},\dots,\mathbf{x}_{N_h}^{h}]^\top \in \mathbb{R}^{N_h\times  d_{\text{node}}}$.

\textbf{Edge Connectivity ($\mathbf{A}^{h}$)}: We use untyped kinematic edges. The edge set $\mathcal{E}_{h}$ is represented by an adjacency matrix $\mathbf{A}^{h} \in \{0,1\}^{N_h \times N_h}$, where $|\mathcal{V}_{h}|=N_h$ matches the hand graph $\mathcal{G}_h$ by construction. $\mathbf{A}^{h}_{ij}=1$ indicates a kinematic connection between nodes $v_i$ and $v_j$. 

\textbf{Global Features ($\mathbf{x}_g^{h}$)}: The global feature vector $\mathbf{x}_g^{h} \in \mathbb{R}^{d_{\text{global}}}$ summarizes object-centric and wrist-level signals:
\begin{equation}
\mathbf{x}_g^{h} = [\Delta \mathbf{p}_{\text{target}}^{h}, \mathbf{v}_{\text{wrist}}^{h}, \boldsymbol{\omega}_{\text{wrist}}^{h}, \mathbf{v}_{\text{obj}}^{h}, \boldsymbol{\omega}_{\text{obj}}^{h}],
\end{equation}
where $\Delta \mathbf{p}_{\text{target}} \in \mathbb{R}^3$ is the displacement from the wrist frame to the object center, $(\mathbf{v}_{\text{wrist}}, \boldsymbol{\omega}_{\text{wrist}})$ are the wrist linear and angular velocities, and $(\mathbf{v}_{\text{obj}}, \boldsymbol{\omega}_{\text{obj}})$ are the object linear and angular velocities in the wrist frame.

\subsubsection{Physical-Property Graph Construction}
\label{subsubsec:physical_property_injection}

The state-graph observation $(\mathbf{X}^{h}_{\text{node}},\, \mathbf{A}^{h},x_g^h)$ captures grasping dynamics and geometry but does not encode embodiment-specific mechanical constraints. To expose these priors to the policy and improve cross-embodiment generalization, we parse the URDF files and build a physical-property graph $\mathcal{G}_{\text{physical}}^{h}$.
Specifically, $\mathcal{G}_{\text{physical}}^{h} = (\mathcal{V}_{\text{physical}}^{h}, \mathcal{E}_{\text{physical}}^{h})$ is constructed to be topology-aligned with the grasping state graph: it shares the same semantic nodes and the same kinematic connectivity, i.e., $|\mathcal{V}_{\text{physical}}^{h}|=|\mathcal{V}_h|=N_h$ and $\mathcal{E}_{\text{physical}}^{h}=\mathcal{E}_h$. Each node $v_j \in \mathcal{V}_{\text{physical}}^{h}$ is characterized by feature vector $\mathbf{x}_j^{\text{physical},h} \in \mathbb{R}^{d_{\text{physical}}}$ that encodes static mechanical priors:
\begin{equation}
\mathbf{x}_j^{\text{physical},h} = [\boldsymbol{\ell}_j^{h}, \mathbf{a}_j^{h}, \mathbf{v}_j^{h}, \boldsymbol{\tau}_j^{h}, \mathbf{l}_j^{h}],
\end{equation}
where $\boldsymbol{\ell}_j \in \mathbb{R}^6$ denotes normalized limits for the three motion primitive axes; $\mathbf{v}_j \in \mathbb{R}^6$ denotes the corresponding normalized velocity bounds in the same motion-primitive axis; $\mathbf{a}_j \in \mathbb{R}^9$ encodes the axis directions in 3D space for the three primitives. $\boldsymbol{\tau}_j \in \mathbb{R}^3$ contains per-axis damping coefficients and $\mathbf{l}_j \in \mathbb{R}^3$ is the link vector from the \emph{parent node} to this node. Stacking all physical node features yields $\mathbf{X}_{\text{physical}}^{h}$:
\begin{equation}
\mathbf{X}_{\text{physical}}^{h} = [\mathbf{x}_1^{\text{physical},h},\dots,\mathbf{x}_{N_h}^{\text{physical},h}]^\top \in \mathbb{R}^{N_h\times d_{\text{physical}}}.
\end{equation}

We encode node features $\mathbf{x}_j^{\text{physical},h}$ into a learnable physical embedding in Section~\ref{subsec:magcn_design}.

\textbf{Activation Mask $\mathbf{M}_{\text{activation}}^{h}$}: Not all nodes support all motion-primitives due to mechanical constraints. To encode which primitives are physically realizable per node, we derive an activation mask aligned with the motion-primitive space in Section~\ref{subsubsec:primitive_action_space}. For each node $v_i \in \mathcal{V}_h$, we define a vector
\begin{equation}
\mathbf{m}_i^{h} = [m_{i,\text{FLEX}}^{h},\, m_{i,\text{ABD}}^{h},\, m_{i,\text{ROT}}^{h}]^\top \in \{0,1\}^3,
\end{equation}

where $m_{i,p}^{h}=1$ if primitive $p\in\{\text{FLEX},\text{ABD},\text{ROT}\}$ is physically realizable at node $i$ for hand $h$, and $m_{i,p}^{h}=0$ otherwise. Stacking all nodes gives
\begin{equation}
\mathbf{M}_{\text{activation}}^h = [\mathbf{m}_1^{h},\dots,\mathbf{m}_{N_h}^{h}]^\top \in \{0,1\}^{N_h\times 3}.
\end{equation}
Equivalently, $\mathbf{M}_{\text{activation}}^h$ can be derived from the indexing rule $\mathcal{M}_h$ by setting $m_{i,p}^{h}=1$ if there exists a DoF index $j\in\{1,\dots,L_h\}$ such that $\text{node}(j)=i$ and $\text{prim}(j)=p$.
We use $\mathbf{M}_{\text{activation}}^h$ to (i) condition the decoder on which primitive axes are executable, and (ii) define an action-feasibility penalty term $r_{\text{pen}}$ in the reward (Section~\ref{subsec:reward}) to discourage non-executable outputs on inactive primitive axes.

\subsubsection{MAGCN Model Design}
\label{subsec:magcn_design}

DexGrasp-Zero is the overall shared policy $\pi_\theta$ introduced in Eq.~\eqref{eq:zero_control_loop}. In our implementation, $\pi_\theta$ is parameterized by the \textit{Morphology-Aligned Graph Convolutional Network} (MAGCN) with three modules shown in Fig.~\ref{fig:overview}: Morphology-Aligned Graph Encoding, Physical Property Encoding, and Decoding. To keep the notation concise, we name the modules used in these modules as functions: a node-wise physical encoder $\phi_{\text{p}}$, a global encoder $\phi_{\text{g}}$, a node-embedding encoder $\phi_{\text{node}}$, a node-action decoder $\phi_{\text{dec,node}}$, and a wrist-action decoder $\phi_{\text{dec,wrist}}$, as shown in Fig.~\ref{fig:overview}.

\paragraph{Morphology-Aligned Graph Encoding Module}
We first encode the grasping observation graph (Section~\ref{subsubsec:Graph-based State Representation}) using a GCN backbone to capture kinematic topology and propagate local signals along the hand graph. Concretely, the MAGCN encoder produces (i) a global embedding $\mathbf{E}_{\text{g}}^{h}$ from global features $\mathbf{x}_g^h$, and (ii) a node-embedding matrix $\mathbf{E}_{\text{node}}^{h}$ from the node-feature matrix $\mathbf{X}_{\text{node}}^{h}$ and adjacency $\mathbf{A}^h$:
\begin{equation}
\mathbf{E}_{\text{g}}^{h} = \phi_{\text{g}}\big(\mathbf{x}_{g}^{h}\big),
\qquad
\mathbf{E}_{\text{node}}^{h} = \phi_{\text{node}}\big(\mathbf{X}_{\text{node}}^{h},\, \mathbf{A}^h\big).
\label{eq:magcn_encoder_abstract}
\end{equation}

\paragraph{Physical Property Encoding Module}
In order to inject embodiment-specific physical information (Section~\ref{subsubsec:physical_property_injection}) and enable the policy to adaptively compensate for varying link lengths and actuation limits thereby ensuring precise and stable grasping, we encode the physical-property graph $\mathcal{G}_{\text{physical}}^{h}$ with a node-wise MLP $\phi_{\text{p}}$ to a learnable embedding. For each node $j=1,\dots,N_h$,
\begin{equation}
\mathbf{e}_{j}^{\text{p},h} = \phi_{\text{p}}\big(\mathbf{x}_j^{\text{physical},h}\big) \in \mathbb{R}^{d_p},
\end{equation}
and stacking yields the physical embedding matrix
\begin{equation}
\mathbf{E}_{\text{p}}^{h} = \big[\mathbf{e}_{1}^{\text{p},h},\dots,\mathbf{e}_{N_h}^{\text{p},h}\big]^\top \in \mathbb{R}^{N_h\times d_p}.
\end{equation}
% $\mathbf{E}_{\text{p}}^{h}$ is a hand-specific prior derived from the hand description and reused across timesteps.

\paragraph{Layer-wise Physical Injection in MAGCN}
With the physical information encoded in $\mathbf{E}_{\text{p}}^{h}$, we can inject these embodiment-specific priors into the state-graph encoder by fusing $\mathbf{E}_{\text{p}}^{h}$ at every GCN layer~\cite{GCN}. Specifically, we instantiate $\phi_{\text{node}}$ as $L$ layers of GCN with layer-wise physical injection.
Equivalently, the node encoder can be written as
\begin{equation}
\mathbf{E}_{\text{node}}^{h} = \phi_{\text{node}}\big(\mathbf{X}_{\text{node}}^{h},\, \mathbf{A}^h,\, \mathbf{E}_{\text{p}}^h\big),
\end{equation}
and is computed as follows. At each layer, we concatenate the current node representation with the physical prior and perform message passing with normalized adjacency:
\begin{align}
&\mathbf{H}^{h,(0)} = \mathbf{X}^{h}_{\text{node}}, \\
&\mathbf{Z}^{h,(l)} = concat(\mathbf{H}^{h,(l-1)}, \mathbf{E}_{\text{p}}^{h}), \\
&\mathbf{H}^{h,(l)} = \sigma\big(\text{Ln}(\hat{\mathbf{A}}^{h}\, \mathbf{Z}^{h,(l)} \mathbf{W}^{(l)})\big), \quad l=1,\dots,L, \\
&\mathbf{E}_{\text{node}}^{h}=\mathbf{H}^{h,(L)},
\end{align}
where $\text{Ln}$ denotes LayerNorm and $\hat{\mathbf{A}}^{h}$ is the standard GCN normalized adjacency with self-loops: $\hat{\mathbf{A}}^{h}=(\tilde{\mathbf{D}}^{h})^{-\frac{1}{2}}\tilde{\mathbf{A}}^{h}(\tilde{\mathbf{D}}^{h})^{-\frac{1}{2}}$, $\tilde{\mathbf{A}}^{h}=\mathbf{A}^{h}+\mathbf{I}$, and $\tilde{\mathbf{D}}^{h}_{ii} = \sum_j \tilde{\mathbf{A}}^{h}_{ij}$.

\paragraph{Decoding Module}
This module jointly parameterizes the motion-primitive space (Section~\ref{subsubsec:primitive_action_space}) using two decoders (Fig.~\ref{fig:overview}c). For node-level primitives, we condition on the activation mask by concatenating the node embedding with the corresponding mask row and decoding with $\phi_{\text{dec,node}}$:
\begin{equation}
\tilde{\mathbf{e}}_{i}^{\text{node},h} = \text{concat}(\mathbf{E}_{\text{node}}^{h}[i], \mathbf{M}_{\text{activation}}^{h}[i]),
\end{equation}
\begin{equation}
\boldsymbol{\alpha}_{i}^{h} = \phi_{\text{dec,node}}\big(\tilde{\mathbf{e}}_{i}^{\text{node},h}\big) \in \mathbb{R}^3, \quad i=1,\dots,N_h.
\end{equation}
For the wrist 6-DoF command, we decode from the global embedding $\mathbf{E}_{\text{g}}^{h}$ and the wrist node embedding $\mathbf{e}_{\text{wrist}}^{h}$ :
\begin{equation}
(\Delta \mathbf{p}_{w}^{h},\, \Delta \boldsymbol{\theta}_{w}^{h}) = \phi_{\text{dec,wrist}}\big( \text{concat}(\mathbf{E}_{\text{g}}^{h} , \mathbf{e}_{\text{wrist}}^{h})\big),
\end{equation}
where $\mathbf{e}_{\text{wrist}}^{h}$ denotes the embedding of the wrist node in $\mathbf{E}_{\text{node}}^{h}$.
Finally, stacking wrist and node-level outputs yields $\boldsymbol{\alpha}_{\text{prim}}^{h}$ as defined in Eq.~\eqref{eq:alpha_prim_stack}.

\subsubsection{Reward design}
\label{subsec:reward}
Following RobustDexGrasp~\cite{robustdexgrasp}, we define a grasping reward $r_{\text{grasp}}$ that encourages stable contact formation under geometric and physical constraints, and add an additional feasibility penalty $r_{\text{pen}}$ tailored for our cross-embodiment motion-primitive space:
\begin{equation}
r = r_{\text{grasp}} + r_{\text{pen}}.
\end{equation}
\textbf{Grasping reward:} $ r_{\text{grasp}}=w_{\text{dis}}\,r_{\text{dis}}+w_{\text{contact}}\,r_{\text{contact}}+w_{\text{force}}\,r_{\text{force}}+w_{\text{reg}}\,r_{\text{reg}}$, where $r_{\text{dis}}=-\sum_{i=1}^{N_h}\|\mathbf{d}_{i}^{h}\|_2$ penalizes distances from hand nodes to the object surface; $ r_{\text{contact}}=\sum_{i=1}^{N_h}c_{i}^{h}$ rewards contact formation; $r_{\text{force}}=-\sum_{i=1}^{N_h}\max(0,f_{i}^{h}-f_0)^2$ penalizes excessive contact forces above threshold $ f_0$; $r_{\text{reg}}=-\|\Delta \mathbf{q}^{h}\|_2$ regularizes joint command increments.

\noindent\textbf{Feasibility penalty:} $r_{\text{pen}} = -w_{\text{pen}} \sum_{i=1}^{N_h} \| (\mathbf{1}-\mathbf{m}_i^{h}) \odot \boldsymbol{\alpha}_{i}^{h} \|_2^2 $ suppresses outputs on inactive primitive axes.

All weights ($w_{\text{dis}}, w_{\text{contact}}, w_{\text{force}}, w_{\text{reg}}, w_{\text{pen}} > 0$) are shared across embodiments to ensure consistent learning objectives.

\subsection{Sim-to-Real Transfer via Privileged Distillation}
\label{sec:sim2real}
% In real-world deployment, contact states and interaction forces are typically not directly observable, while they are readily available as privileged signals in simulation. We follow the teacher--student distillation strategy of RobustDexGrasp~\cite{robustdexgrasp}: we first train a privileged \emph{teacher} policy in simulation with access to contact/force-related observations, and then distill it into a \emph{student} policy that operates without such privileged inputs. The student use MAGCN as backbone and is equipped with an LSTM to perform temporal state estimation, enabling it to implicitly recover missing information (e.g., contacts and forces) from observation histories and to execute the distilled grasping behavior in real scenes. Detail is illustrated in the Supplementary Material.
In real-world deployment, contact states and interaction forces are typically not directly observable, while they are readily available as privileged signals in simulation. We follow the teacher--student distillation strategy of RobustDexGrasp~\cite{robustdexgrasp}: we first train a privileged \emph{teacher} policy in simulation with access to contact/force-related observations, and then distill it into a \emph{student} policy that operates without such privileged inputs. The student uses MAGCN as the backbone and is equipped with an LSTM to perform temporal state estimation, enabling it to implicitly recover missing information (e.g., contacts and forces) from observation histories and to execute the distilled grasping behavior in real scenes (see Supplementary Sec.~\ref{subsubsec:supp_distill_details} and Sec.~\ref{subsubsec:supp_hw_details}).

\section{Experiments}
\label{sec:experiments}

\begin{table*}[ht!]
\centering
% \caption{Cross-embodiment training and zero-shot transfer results (success rate). Variants: \textit{w/o $\mathcal{G}_{\text{physical}}$ priors}, \textit{early fusion} (input concatenation), \textit{w/o motion primitives} (raw joint commands), \textit{w/o $M_{\text{activation}}$ \& $r_{\text{pen}}$}, and \textbf{full model}.}
\caption{Cross-embodiment training and zero-shot transfer results (success rate). 
Variants of our method: 
\textit{w/o  $ \mathcal{G}_{\text{physical}} $  priors} removes hand-specific physical property encoding; 
\textit{early fusion} concatenates physical features with node states at input (instead of layer-wise fusion in GCN); 
\textit{w/o motion primitives} replaces the  motion-primitive space with raw joint commands; 
\textit{w/o  $ M_{\text{activation}} \& r_{\text{pen}} $ } disables the activation mask conditioning and the corresponding action-feasibility penalty in the reward; 
\textbf{full model} is our complete DexGrasp-Zero policy.}
\label{tab:cross}
\small
\setlength{\tabcolsep}{4.5pt}
\renewcommand{\arraystretch}{1.1}
\begin{tabular}{llcccc|cc|cc}
\hline
\multirow{2}{*}{\textbf{Method}}& \multirow{2}{*}{\textbf{Variant}}& \multicolumn{4}{c}{\textbf{Training Hands}} & \multicolumn{2}{c}{\textbf{Unseen Hands}} & \multicolumn{2}{c}{\textbf{Average}} \\
\cline{3-6} \cline{7-8} \cline{9-10}
 & & Allegro & Shadow & Ability & Schunk & LEAP & Inspire & Seen & Unseen \\
\hline
CrossDex~\cite{cross} & per-object & 0.81 & 0.85 & \textbf{0.90} & 0.90 & 0.34 & 0.44 & 0.865 & 0.39 \\
\hline
CrossDex~\cite{cross} & multi-object & 0.39 & 0.69 & 0.42 & 0.60 & 0.19 & 0.34 & 0.525 & 0.265 \\
\cdashline{1-10}
\multirow{5}{*}{DexGrasp-Zero (Ours)}
& \textit{w/o motion primitives} & 0.52 & 0.51 & 0.64 & 0.59 & 0.39 & 0.29 & 0.565 & 0.34 \\
& \textit{early fusion} & 0.42 & 0.47 & 0.59 & 0.54 & 0.46 & 0.34 & 0.505 & 0.40 \\
& \textit{w/o  $ \mathcal{G}_{\text{physical}} $  priors} & 0.91 & 0.89 & \textbf{0.90} & 0.84 & 0.82 & 0.79 & 0.885 & 0.805 \\
& \textit{w/o  $ M_{\text{activation}} \& r_{\text{pen}} $ } & \textbf{0.92} & 0.91 & \textbf{0.90} & 0.81 & 0.50 & 0.76 & 0.885 & 0.63 \\
& \textbf{full model} & \textbf{0.92} & \textbf{0.95} & \textbf{0.90} & \textbf{0.91} & \textbf{0.93} & \textbf{0.82} & \textbf{0.92} & \textbf{0.85} \\
\hline
\end{tabular}
\vspace{-1em}
\end{table*}

% We evaluate DexGrasp-Zero in simulation and real robots. In simulation, we benchmark cross-embodiment learning with zero-shot transfer across six dexterous hands on the YCB object split, and further analyze single-hand transfer and ablations. Key observations are: (i) a single cross-hand policy achieves strong zero-shot performance on unseen hands, (ii) physical-property priors and motion primitives are both critical for robust transfer, and (iii) the simulation-trained policy achieves performance close to an intra-hand oracle on three real robot platforms.

We evaluate DexGrasp-Zero in both simulation and real-world, including cross-hand transfer, single-hand transfer, and ablations.

\subsection{Experimental Setup}

\subsubsection{Simulation}
We evaluate DexGrasp-Zero on six dexterous robot hands. Following the cross-embodiment protocol used by CrossDex~\cite{cross}, we train a single policy on four seen hands (Allegro, Shadow, Ability, and Schunk) and evaluate zero-shot transfer on two unseen hands (LEAP\citep{leaphand} and Inspire) in the RaiSim simulator~\cite{raisim}.

\paragraph{Dataset and Benchmark}~

\noindent\textbf{CrossDex/YCB benchmark.} To benchmark cross-hand generalization, we adopt the 45-object YCB split from CrossDex~\cite{cross} where all objects are used for both training and testing. This protocol keeps the object set fixed so performance differences primarily reflect embodiment variation rather than changes in objects. We compare against CrossDex\citep{cross} under (i) its original \emph{per-object} setting (one policy trained per object, averaged across objects) and (ii) a \emph{multi-object} adaptation (a single policy jointly trained on all objects).

\noindent\textbf{GraspXL benchmark.} To study single-hand training and cross-hand transfer under a standard single-hand benchmark, we follow the protocol of GraspXL\citep{graspxl}: the training set consists of 26 ShapeNet~\cite{chang2015shapenetinformationrich3dmodel} objects and 32 PartNet~\cite{Mo2018PartNetAL} objects, and evaluation is performed on 48 PartNet test objects. We compare against GraspXL in the single-hand setting.

\paragraph{Evaluation Protocol and Metric} In all simulation experiments, we evaluate with 25 grasp trials per object; the object is spawned at a fixed initial position while the hand starts from a randomized initial pose. We report the success rate (SR): the object is lifted up by 0.5~m and held for 2~s without dropping.

\subsubsection{Real-World Evaluation}
\label{sec:real_world_evaluation}
\begin{figure}[t!]
    \centering
    \includegraphics[width=1\linewidth,trim=0 8.5cm 7cm 0cm,clip]{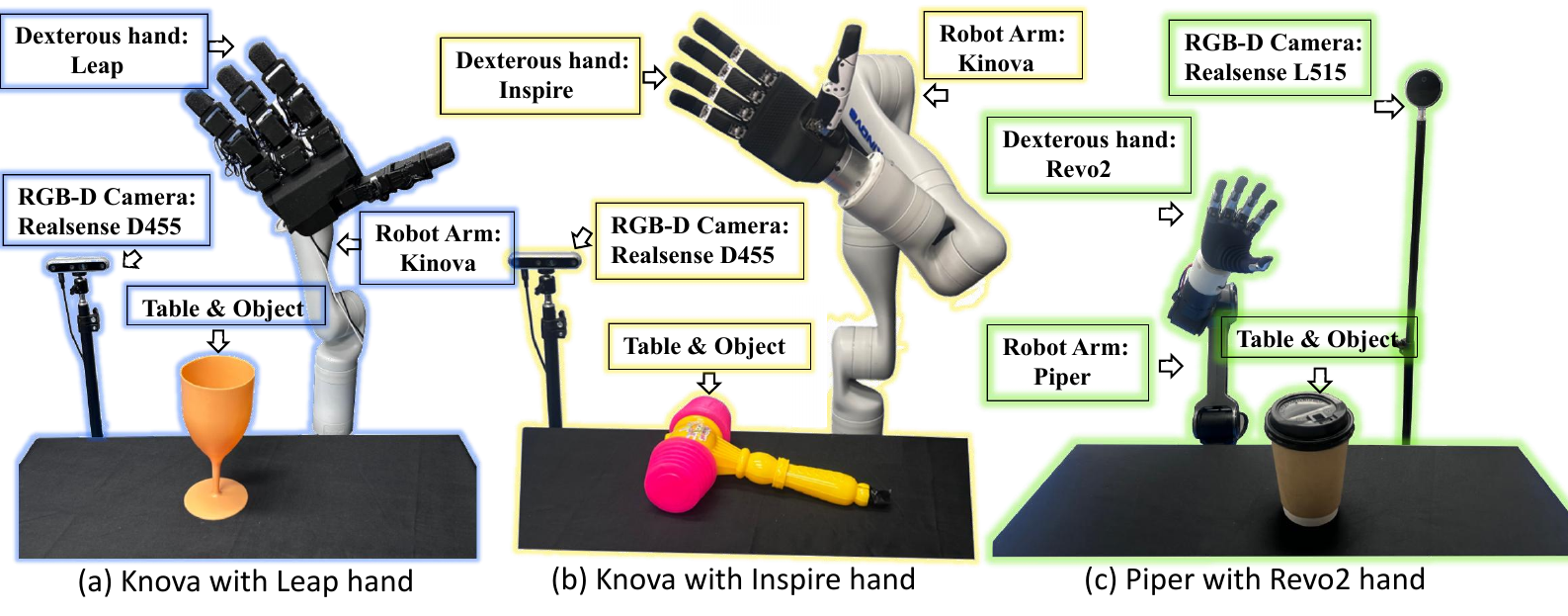}
    \vspace{-1.5em}
    \caption{Hardware setup. We evaluate our method on three robot platforms: (a)Kinova arm with LEAP hand, (b)Kinova arm with Inspire hand, and (c)Piper arm with Revo2 hand.}
    \label{fig:hardware_setup}
    \vspace{-1.3em}
\end{figure}

We deploy distilled policy (Sec. \ref{sec:sim2real}) on 3 robot platforms: (i) Kinova arm with LEAP hand, (ii) Kinova arm with Inspire hand, and (iii) Piper arm with Revo2 hand (Fig.~\ref{fig:hardware_setup}). We evaluate on 10 unseen objects with 5 random poses per object (50 trials per platform). A grasp is successful if the object is lifted to 30~cm and held for 5~s.

\subsection{Implementation Details}

\noindent\textbf{Simulation training.} We train all policies with PPO~\cite{ppo} and on a NVIDIA RTX 3090 GPU for 6000 rounds. The MAGCN backbone uses a 10-layer GCN. We instantiate 3 parallel simulation environment for each (hand, object) pair. 
The simulator runs at a 20~Hz. Each episode has 120 exploration steps, followed by 30 steps with an additional lift signal. We use the following reward coefficients: $w_{\text{dis}}=0.3, w_{\text{contact}}=1.0, w_{\text{force}}=0.5, w_{\text{reg}}=1.5, w_{\text{pen}}=0.3$.

\noindent\textbf{Real-world deployment details.} Inference runs on a single NVIDIA RTX 3090 GPU at 20~Hz. Perception is provided by a single RGB-D camera. We standardize the initial robot configuration across platforms: the thumb starts with maximum opening distance. The arm starts 25~cm to the left of the object with the palm facing the object. During execution, we run the grasp controller for 130 steps and then send a lift command to raise the object.

\subsection{Cross-Hand Zero-Shot Transfer from Multi-Hand Training}
To evaluate cross-embodiment generalization, we train a single policy jointly on four hands (Allegro, Shadow, Ability, Schunk) on the 45-object YCB split and evaluate on both the training and unseen hands (LEAP and Inspire) without fine-tuning. We compare against CrossDex~\cite{cross}, a state-of-the-art method with open code for zero-shot cross-embodiment dexterous grasping. %; while She~et~al.~\cite{she2024rl} report higher numbers, their approach is not reproducible due to missing implementation details.
\begin{figure}[t!]
    \centering
    \includegraphics[width=0.95\linewidth,trim=0.3cm 12.2cm 23cm 0cm,clip]{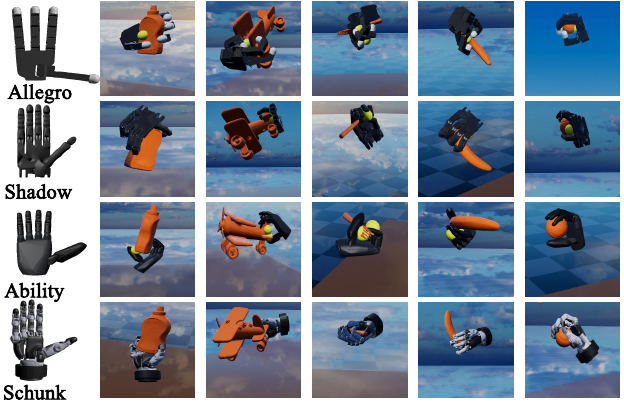}
    \caption{Simulated grasps of training hands on 5 diverse objects. }
    \label{fig:sim_grasps}
    \vspace{-1em}
\end{figure}

As shown in Table~\ref{tab:cross}, DexGrasp-Zero achieves strong performance across both seen and unseen hand embodiments. In particular, our full model reaches \textbf{92.0\%} average success on the four training hands and \textbf{85.0\%} on two unseen hands. Compared with the CrossDex multi-object baseline (52.5\% seen/26.5\% unseen), we substantially improve zero-shot transfer on unseen hands (26.5\% $\rightarrow$ 85\%), driven by our morphology-aligned graph design, physical-property injection, and action-space alignment.

\subsection{Cross-Hand Zero-Shot Transfer from Single-Hand Training}
Here we ask a more practical question: if we train DexGrasp-Zero on a \emph{single} hand, can the resulting policy still transfer well to other embodiments, and how competitive is it compared with methods tailored for single-hand training?

Using the GraspXL benchmark, we train DexGrasp-Zero on one embodiment at a time and evaluate the resulting policy on all hands without any adaptation. We report success rates in Table~\ref{tab:single_hand_transfer} together with the GraspXL, a method design for single-hand training and can not transfer to new hand.

\begin{table}[t]
\centering
\caption{Single-hand training transfer on PartNet (success rate): rows show the training embodiment for DexGrasp-Zero and columns show the test embodiment. Entries marked with $^{\dag}$ indicate \emph{in-domain} evaluation.}
\label{tab:single_hand_transfer}
\tiny
{\renewcommand{\arraystretch}{1.08}
\setlength{\tabcolsep}{1.5pt}
\resizebox{\linewidth}{!}{%
\begin{tabular}{@{}l@{\hspace{6pt}}lcccccc@{}}
\hline
\textbf{Method} & \textbf{Train Hand} & \textbf{Allegro} & \textbf{Shadow} & \textbf{Ability} & \textbf{Schunk} & \textbf{LEAP} & \textbf{Inspire} \\
\hline
GraspXL~\cite{graspxl} & -- & 0.94 & 0.93 & 0.91 & 0.90 & 0.95 & 0.91 \\
\hline
\multirow{6}{*}{DexGrasp-Zero (Ours)}
& Allegro & $0.93^{\dag}$ & 0.83 & 0.80 & 0.69 & 0.48 & 0.77 \\
& Shadow  & 0.55 & $0.98^{\dag}$ & 0.80 & 0.82 & 0.77 & 0.94 \\
& Ability & 0.60 & 0.60 & $0.86^{\dag}$ & 0.86 & 0.88 & 0.90 \\
& Schunk  & 0.61 & 0.52 & 0.83 & $0.92^{\dag}$ & 0.90 & 0.94 \\
& LEAP    & 0.50 & 0.68 & 0.65 & 0.69 & $0.90^{\dag}$ & 0.69 \\
& Inspire & 0.61 & 0.83 & 0.87 & 0.88 & 0.88 & $0.96^{\dag}$ \\
\hline
\end{tabular}}}
\vspace{-2.0em}
\end{table}
Overall, single-hand training with DexGrasp-Zero also achieves strong zero-shot transfer. Policies trained on Shadow or Schunk reach \textbf{0.94 on unseen Inspire}, surpassing the GraspXL specialist (0.91). Training on Schunk also yields \textbf{0.90 on LEAP} and exceeds GraspXL on Schunk itself (0.92 vs.\ 0.90), demonstrating effective cross-embodiment generalization. But this single-hand transfer performance is not uniformly stable, it depends heavily on morphological and dynamic similarities between the source and target hands.

\subsection{Ablation Studies}
We conduct ablation studies to quantify the contributions of four key design choices: (1)conditioning the policy on URDF-derived \emph{physical-property} priors via $\mathcal{G}_{\text{physical}}$, (2)the proposed layer-wise injection of task features with physical priors (vs. early fusion), (3)the hand-agnostic motion-primitive space, and (4)the activation mask $M_\text{activation}$ conditioning with the corresponding action-feasibility penalty $r_\text{pen}$.
Result is summarized in Table~\ref{tab:cross}, removing any component degrades performance. Specifically:
\begin{itemize}
    \item \textbf{Without motion primitives}, performance degrades markedly: the average success rate decreases from 92.0\% to 56.5\% on seen hands and from 85.0\% to 34.0\% on unseen hands, confirming that the motion-primitive provides essential inductive bias that facilitates structural and semantic alignment across different morphologies.
    \item \textbf{With early fusion ($\mathcal{G}_{\text{task}}\oplus\mathcal{G}_{\text{physical}}$)}: replacing layer-wise injection with a one-shot feature concatenation performs substantially worse (from 92.5\% to 50.0\% on seen hands) and exhibits \emph{highly unstable} learning dynamics, suggesting that injecting physical priors at every layer is important for stable optimization.
    \item \textbf{Without $\mathcal{G}_{\text{physical}}$ priors}, performance drops from 92.0\% to 88.5\% on seen hands and from 85.0\% to 80.5\% on unseen hands. This indicates that URDF-derived physical constraints provide useful embodiment conditioning, especially for zero-shot transfer.
    \item \textbf{Without $M_{\text{activation}}$ and $r_{\text{pen}}$}, performance drops from 85.0\% to 63.0\% on unseen hands (while remaining relatively high on seen hands: 92.0\% to 88.5\%). This suggests that explicitly conditioning on executable primitive axes and penalizing infeasible outputs is crucial for preventing invalid actions that do not transfer across embodiments.

\end{itemize}
These results demonstrate that physical-property conditioning, a stable injection strategy, motion primitives, and feasibility-aware action conditioning are all important for achieving strong zero-shot generalization across diverse hand embodiments.

\subsection{ Real-World Deployment}
Under the unified experimental setup described in Section \ref{sec:real_world_evaluation}, we evaluate our policy on three robot platforms.

%\subsubsection{Result Analysis}
We evaluate real-world transfer performance under the unified setup by comparing our cross-hand policy with an intra-hand oracle and an ablation without $\mathcal{G}_{\text{physical}}$ (Table~\ref{tab:cross_hand_analysis}). This ablation is chosen for real-world testing because physical priors are essential for embodiment realism.

\begin{table}[t!]
\centering
\caption{Real-world zero-shot grasping success rate on unseen dexterous hands.}
\label{tab:cross_hand_analysis}
\begin{tabular}{lcccc}
\hline
\textbf{Method} & \textbf{LEAP} & \textbf{Inspire} & \textbf{Revo2} & \textbf{Average}\\
\hline
Intra-hand (Oracle) & 0.90 & 0.90 & 0.78 & 0.86 \\
Cross-hand w/o $\mathcal{G}_{\text{physical}}$ & 0.84 & 0.80 & 0.62 & 0.75\\
Cross-hand (Ours)   & 0.88 & 0.86 & 0.72 & 0.82\\
\hline
\end{tabular}
\vspace{-1em}
\end{table}

The results in Table~\ref{tab:cross_hand_analysis} reveal two key insights. First, our cross-hand policy achieves performance close to the intra-hand oracle (e.g., 90\% vs. 88\% on LEAP), demonstrating that training on diverse hands enables strong zero-shot transfer without sacrificing much capability. Second, removing URDF-derived physical-property priors causes a consistent performance drop (e.g., from 0.88 to 0.84 on LEAP and from 0.72 to 0.62 on Revo2), confirming that explicit physical constraints encoded by $\mathcal{G}_{\text{physical}}$ are essential for effective cross-embodiment generalization and for adaptively compensating varying link lengths and actuation limits. Revo2 shows the lowest successrate, likely due to its smaller size, which makes large objects harder to stably grasp.

\section{Conclusion}
\label{sec:conclusion}
% We presented \textbf{DexGrasp-Zero} for zero-shot cross-embodiment dexterous grasping. Our main contributions are a morphology-aligned graph representation with motion primitives for structural/semantic alignment across morphologies, and a Physical Property Injection module that incorporates URDF-derived constraints for stable grasping.
We presented \textbf{DexGrasp-Zero} for zero-shot cross-embodiment dexterous grasping. Our main contributions are a morphology-aligned graph representation with motion primitives for structural and control-semantic alignment across morphologies and MAGCN, a GCN policy network with Physical Property Injection that incorporates URDF-derived constraints for stable grasping.
Experiments show strong transfer to unseen hands (85\% in simulation; 82\% on real robots), validating its potential as a universal paradigm for cross-embodiment dexterous manipulation.

\bibliographystyle{plainnat}

\bibliography{references}

@INPROCEEDINGS{YCB,
  author={Calli, Berk and Singh, Arjun and Walsman, Aaron and Srinivasa, Siddhartha and Abbeel, Pieter and Dollar, Aaron M.},
  booktitle={International Conference on Advanced Robotics (ICAR)}, 
  title={\href{https://ieeexplore.ieee.org/stamp/stamp.jsp?tp=&arnumber=7251504}{The YCB object and Model set: Towards common benchmarks for manipulation research}}, 
  year={2015},
  volume={},
  number={},
  pages={510-517},
  doi={10.1109/ICAR.2015.7251504},
  url={https://ieeexplore.ieee.org/stamp/stamp.jsp?tp=&arnumber=7251504}
}

@article{leaphand,
      title={\href{https://www.roboticsproceedings.org/rss19/p089.pdf}{LEAP Hand: Low-Cost, Efficient, and Anthropomorphic Hand for Robot Learning}},
      author={Shaw, Kenneth and Agarwal, Ananye and Pathak, Deepak},
      journal={Robotics: Science and Systems (RSS)},
      year={2023},
      doi={10.15607/RSS.2023.XIX.089},
      url={https://www.roboticsproceedings.org/rss19/p089.pdf}
}

@misc{zhou2025designadaptivemodularanthropomorphic,
      title={\href{https://arxiv.org/pdf/2511.22100}{Design of an Adaptive Modular Anthropomorphic Dexterous Hand for Human-like Manipulation}}, 
      author={Zelong Zhou and Wenrui Chen and Zeyun Hu and Qiang Diao and Qixin Gao and Yaonan Wang},
      year={2025},
      url={https://arxiv.org/abs/2511.22100}, 
}

@article{Santello1998PosturalHS,
  title={\href{https://www.jneurosci.org/content/18/23/10105}{Postural Hand Synergies for Tool Use}},
  author={Marco Santello and Martha Flanders and John F. Soechting},
  journal={The Journal of Neuroscience},
  year={1998},
  volume={18},
  pages={10105 - 10115},
  url={https://www.jneurosci.org/content/18/23/10105}
}

@misc{weng2025bidexhanddesignevaluationopensource,
      title={\href{{https://arxiv.org/abs/2504.14712}}{BiDexHand: Design and Evaluation of an Open-Source 16-DoF Biomimetic Dexterous Hand}}, 
      author={Zhengyang Kris Weng},
      year={2025},
      url={https://arxiv.org/abs/2504.14712}, 
}

@book{Neumann2009KinesiologyOT,
    author ={Donald A. Neumann} ,
    title ={Kinesiology of the Musculoskeletal System : Foundations for Rehabilitation} ,
    publisher = {Elsevier Health Sciences},
    year = {2009}
}

@INPROCEEDINGS{dro,
  author={Wei, Zhenyu and Xu, Zhixuan and Guo, Jingxiang and Hou, Yiwen and Gao, Chongkai and Cai, Zhehao and Luo, Jiayu and Shao, Lin},
  booktitle={IEEE International Conference on Robotics and Automation (ICRA)}, 
  title={\href{https://ieeexplore.ieee.org/stamp/stamp.jsp?tp=&arnumber=11127754}{$\mathcal{D}(\mathcal{R}, \mathcal{O})$ Grasp: A Unified Representation of Robot and Object Interaction for Cross-Embodiment Dexterous Grasping}}, 
  year={2025},
  volume={},
  number={},
  pages={4982-4988},
  doi={10.1109/ICRA55743.2025.11127754},
  url={https://ieeexplore.ieee.org/stamp/stamp.jsp?tp=&arnumber=11127754}
}

@article{tro,
  title={\href{https://arxiv.org/pdf/2510.12724}{$\mathcal{T}(\mathcal{R}, \mathcal{O})$ Grasp: Efficient Graph Diffusion of Robot-Object Spatial Transformation for Cross-Embodiment Dexterous Grasping}},
  author={Fei, Xin and Xu, Zhixuan and Fang, Huaicong and Zhang, Tianrui and Shao, Lin},
  journal={arXiv preprint arXiv:2510.12724},
  year={2025},
  url={https://arxiv.org/pdf/2510.12724}
}

@inproceedings{24eccv,
  title={\href{https://www.ecva.net/papers/eccv_2024/papers_ECCV/papers/04377.pdf}{Learning Cross-Hand Policies of High-DOF Reaching and Grasping}},
  author={Qijin She and Shishun Zhang and Yunfan Ye and Ruizhen Hu and Kai Xu},
  booktitle={European Conference on Computer Vision (ECCV)},
  year={2024},
  url={https://www.ecva.net/papers/eccv_2024/papers_ECCV/papers/04377.pdf}
}

@inproceedings{cross,
  author = {Yuan, Haoqi and Zhou, Bohan and Fu, Yuhui and Lu, Zongqing},
  booktitle = {International Conference on Learning Representations (ICLR)},
  pages = {81413--81434},
  title = {\href{https://proceedings.iclr.cc/paper_files/paper/2025/file/ca8c6f28d8ba1e732e3f217ab05c4ec0-Paper-Conference.pdf}{Cross-Embodiment Dexterous Grasping with Reinforcement Learning}},
  url = {https://proceedings.iclr.cc/paper_files/paper/2025/file/ca8c6f28d8ba1e732e3f217ab05c4ec0-Paper-Conference.pdf},
  volume = {2025},
  year = {2025}
}

@INPROCEEDINGS{getzero,
  author={Patel, Austin and Song, Shuran},
  booktitle={IEEE International Conference on Robotics and Automation (ICRA)}, 
  title={\href{https://ieeexplore.ieee.org/stamp/stamp.jsp?tp=&arnumber=11127922}{GET-Zero: Graph Embodiment Transformer for Zero-Shot Embodiment Generalization}}, 
  year={2025},
  volume={},
  number={},
  pages={14262-14269},
  doi={10.1109/ICRA55743.2025.11127922},
  url={https://ieeexplore.ieee.org/stamp/stamp.jsp?tp=&arnumber=11127922}
}

@misc{GEOMatch++,
  title={\href{https://arxiv.org/abs/2412.18998}{GeoMatch++: Morphology Conditioned Geometry Matching for Multi-Embodiment Grasping}},
  author={Yunze Wei and Maria Attarian and Igor Gilitschenski},
  year={2024},
  eprint={2412.18998},
  archivePrefix={arXiv},
  primaryClass={cs.RO},
  url={https://arxiv.org/abs/2412.18998},
}

@misc{GCN,
      title={\href{https://arxiv.org/abs/1609.02907}{Semi-Supervised Classification with Graph Convolutional Networks}}, 
      author={Thomas N. Kipf and Max Welling},
      booktitle={International Conference on Learning Representations (ICLR)},
      year={2017},
      url={https://arxiv.org/abs/1609.02907}
}

@misc{raisim,
  author={J. Hwangbo and D. Kang},
  title={\href{https://raisim.com/}{Raisim: A High-Performance Physics Engine for Robotics}},
  howpublished={\url{https://raisim.com/}},
  year={2023},
  note={Accessed: 2025-03-15}
}

@misc{PPO,
  title={\href{https://arxiv.org/pdf/1707.06347}{Proximal Policy Optimization Algorithms}},
  author={John Schulman and Filip Wolski and Prafulla Dhariwal and Alec Radford and Oleg Klimov},
  year={2017},
  eprint={1707.06347},
  archivePrefix={arXiv},
  primaryClass={cs.LG},
  url={https://arxiv.org/pdf/1707.06347}
}

@article{unigrasp,
  title={\href{https://doi.org/10.1109/LRA.2020.2969946}{UniGrasp: Learning a Unified Model to Grasp With Multifingered Robotic Hands}},
  volume={5},
  pages={2286--2293},
  number={2},
  journal={IEEE Robotics and Automation Letters (RAL)},
  author={Shao, Lin and Ferreira, Fabio and Jorda, Mikael and Nambiar, Varun and Luo, Jianlan and Solowjow, Eugen and Ojea, Juan Aparicio and Khatib, Oussama and Bohg, Jeannette},
  year={2020},
  doi={10.1109/LRA.2020.2969946},
  url={https://doi.org/10.1109/LRA.2020.2969946}
}

@InProceedings{unidexgrasp++,
    author    = {Wan, Weikang and Geng, Haoran and Liu, Yun and Shan, Zikang and Yang, Yaodong and Yi, Li and Wang, He},
    title     = {\href{https://openaccess.thecvf.com/content/ICCV2023/papers/Wan_UniDexGrasp_Improving_Dexterous_Grasping_Policy_Learning_via_Geometry-Aware_Curriculum_and_ICCV_2023_paper.pdf}{UniDexGrasp++: Improving Dexterous Grasping Policy Learning via Geometry-Aware Curriculum and Iterative Generalist-Specialist Learning}},
    booktitle = {Proceedings of the IEEE/CVF International Conference on Computer Vision (ICCV)},
    month     = {October},
    year      = {2023},
    pages     = {3891-3902},
    url={https://openaccess.thecvf.com/content/ICCV2023/papers/Wan_UniDexGrasp_Improving_Dexterous_Grasping_Policy_Learning_via_Geometry-Aware_Curriculum_and_ICCV_2023_paper.pdf}
}

@inProceedings{graspxl,
  title={\href{https://www.ecva.net/papers/eccv_2024/papers_ECCV/papers/03801.pdf}{{GraspXL}: Generating Grasping Motions for Diverse Objects at Scale}},
  author={Zhang, Hui and Christen, Sammy and Fan, Zicong and Hilliges, Otmar and Song, Jie},
  booktitle={European Conference on Computer Vision (ECCV)},
  year={2024},
  url={https://www.ecva.net/papers/eccv_2024/papers_ECCV/papers/03801.pdf}
}

@inproceedings{unidexgrasp,
  title={\href{https://openaccess.thecvf.com/content/CVPR2023/papers/Xu_UniDexGrasp_Universal_Robotic_Dexterous_Grasping_via_Learning_Diverse_Proposal_Generation_CVPR_2023_paper.pdf}{UniDexGrasp: Universal Robotic Dexterous Grasping via Learning Diverse Proposal Generation and Goal-Conditioned Policy}},
  author={Yinzhen Xu and Weikang Wan and Jialiang Zhang and Haoran Liu and Zikang Shan and Hao Shen and Ruicheng Wang and Haoran Geng and Yijia Weng and Jiayi Chen and Tengyu Liu and Li Yi and He Wang},
  booktitle={IEEE/CVF Conference on Computer Vision and Pattern Recognition (CVPR)},
  year={2023},
  pages={4737-4746},
  url={https://openaccess.thecvf.com/content/CVPR2023/papers/Xu_UniDexGrasp_Universal_Robotic_Dexterous_Grasping_via_Learning_Diverse_Proposal_Generation_CVPR_2023_paper.pdf}
}

@inproceedings{chen2025clutterdexgrasp,
  title={\href{https://arxiv.org/pdf/2506.14317v1}{ClutterDexGrasp: A Sim-to-Real System for General Dexterous Grasping in Cluttered Scenes}},
  author={Chen, Zeyuan and Yan, Qiyang and Chen, Yuanpei and Wu, Tianhao and Zhang, Jiyao and Ding, Zihan and Li, Jinzhou and Yang, Yaodong and Dong, Hao},
  booktitle={Conference on Robot Learning (CORL)},
  year={2025},
  url={https://arxiv.org/pdf/2506.14317v1}
}

@inproceedings{fang2025anydexgrasp,
  title={\href{https://arxiv.org/abs/2502.16420}{AnyDexGrasp: General Dexterous Grasping for Different Hands with Human-level Learning Efficiency}},
  author={Fang, Hao-Shu and Yan, Hengxu and Tang, Zhenyu and Fang, Hongjie and Wang, Chenxi and Lu, Cewu},
  booktitle={ICLR Workshop: Towards Robots with Human-Level Abilities},
  year={2025},
  url={https://arxiv.org/abs/2502.16420}
}

@inProceedings{artigrasp,
  author={Zhang, Hui and Christen, Sammy and Fan, Zicong and Zheng, Luocheng and Hwangbo, Jemin and Song, Jie and Hilliges, Otmar},
  booktitle={International Conference on 3D Vision (3DV)},
  year={2024},
  title={\href{https://doi.org/10.1109/3DV62453.2024.00016}{ArtiGrasp: Physically Plausible Synthesis of Bi-Manual Dexterous Grasping and Articulation}},
  year={2024},
  pages={235--246},
  doi={10.1109/3DV62453.2024.00016},
  url={https://doi.org/10.1109/3DV62453.2024.00016}
}

@inproceedings{Mo2018PartNetAL,
  title={\href{https://openaccess.thecvf.com/content_CVPR_2019/papers/Mo_PartNet_A_Large-Scale_Benchmark_for_Fine-Grained_and_Hierarchical_Part-Level_3D_CVPR_2019_paper.pdf}{PartNet: A Large-Scale Benchmark for Fine-Grained and Hierarchical Part-Level 3D Object Understanding}},
  author={Kaichun Mo and Shilin Zhu and Angel X. Chang and L. Yi and Subarna Tripathi and Leonidas J. Guibas and Hao Su},
  booktitle={IEEE/CVF Conference on Computer Vision and Pattern Recognition (CVPR)},
  year={2018},
  pages={909-918},
  url={https://openaccess.thecvf.com/content_CVPR_2019/papers/Mo_PartNet_A_Large-Scale_Benchmark_for_Fine-Grained_and_Hierarchical_Part-Level_3D_CVPR_2019_paper.pdf}
}

@ARTICLE{FunGrasp,
  author={Huang, Linyi and Zhang, Hui and Wu, Zijian and Christen, Sammy and Song, Jie},
  journal={IEEE Robotics and Automation Letters (RAL)},
  title={\href{https://doi.org/10.1109/LRA.2025.3561573}{FunGrasp: Functional Grasping for Diverse Dexterous Hands}},
  year={2025},
  volume={10},
  number={6},
  pages={6175--6182}, 
  doi={10.1109/LRA.2025.3561573},
  url={https://doi.org/10.1109/LRA.2025.3561573}
}

@misc{chang2015shapenetinformationrich3dmodel,
  title={\href{https://arxiv.org/abs/1512.03012}{ShapeNet: An Information-Rich 3D Model Repository}},
  author={Angel X. Chang and Thomas Funkhouser and Leonidas Guibas and Pat Hanrahan and Qixing Huang and Zimo Li and Silvio Savarese and Manolis Savva and Shuran Song and Hao Su and Jianxiong Xiao and Li Yi and Fisher Yu},
  year={2015},
  eprint={1512.03012},
  archivePrefix={arXiv},
  primaryClass={cs.GR},
  url={https://arxiv.org/abs/1512.03012}
}

@misc{allegro,
  title={\href{https://www.allegrohand.com/}{Allegro robot hand}},
  author={Wonik Robotics},
  url={https://www.allegrohand.com/}
}

@inproceedings{jiang2021hand,
  title={\href{https://openaccess.thecvf.com/content/ICCV2021/papers/Jiang_Hand-Object_Contact_Consistency_Reasoning_for_Human_Grasps_Generation_ICCV_2021_paper.pdf}{Hand-object contact consistency reasoning for human grasps generation}},
  author={Jiang, Hanwen and Liu, Shaowei and Wang, Jiashun and Wang, Xiaolong},
  booktitle={Proceedings of the IEEE/CVF international conference on computer vision (ICCV)},
  pages={11107--11116},
  year={2021},
  url={https://openaccess.thecvf.com/content/ICCV2021/papers/Jiang_Hand-Object_Contact_Consistency_Reasoning_for_Human_Grasps_Generation_ICCV_2021_paper.pdf}
}

@inproceedings{lu2024ugg,
  title={\href{https://www.ecva.net/papers/eccv_2024/papers_ECCV/papers/08453-supp.pdf}{Ugg: Unified generative grasping}},
  author={Lu, Jiaxin and Kang, Hao and Li, Haoxiang and Liu, Bo and Yang, Yiding and Huang, Qixing and Hua, Gang},
  booktitle={European Conference on Computer Vision (ECCV)},
  pages={414--433},
  year={2024},
  organization={Springer},
  url={https://www.ecva.net/papers/eccv_2024/papers_ECCV/papers/08453.pdf}
}

@inproceedings{zhang2024dexgraspnet,
  title={\href{https://arxiv.org/pdf/2410.23004}{Dexgraspnet 2.0: Learning generative dexterous grasping in large-scale synthetic cluttered scenes}},
  author={Zhang, Jialiang and Liu, Haoran and Li, Danshi and Yu, XinQiang and Geng, Haoran and Ding, Yufei and Chen, Jiayi and Wang, He},
  booktitle={Conference on Robot Learning (CORL)},
  year={2024},
  url={https://arxiv.org/pdf/2410.23004}
}

@INPROCEEDINGS{xu2024dexterous,
  author={Xu, Guo-Hao and Wei, Yi-Lin and Zheng, Dian and Wu, Xiao-Ming and Zheng, Wei-Shi},
  booktitle={Proceedings of the Computer Vision and Pattern Recognition Conference (CVPR)},
  title={\href{https://doi.org/10.1109/CVPR52733.2024.01698}{Dexterous Grasp Transformer}},
  year={2024},
  volume={},
  number={},
  pages={17933--17942}, 
  doi={10.1109/CVPR52733.2024.01698},
  url={https://doi.org/10.1109/CVPR52733.2024.01698}
}

@inproceedings{wang2025unigrasptransformer,
  title={\href{https://openaccess.thecvf.com/content/CVPR2025/supplemental/Wang_UniGraspTransformer_Simplified_Policy_CVPR_2025_supplemental.pdf}{Unigrasptransformer: Simplified policy distillation for scalable dexterous robotic grasping}},
  author={Wang, Wenbo and Wei, Fangyun and Zhou, Lei and Chen, Xi and Luo, Lin and Yi, Xiaohan and Zhang, Yizhong and Liang, Yaobo and Xu, Chang and Lu, Yan and others},
  booktitle={Proceedings of the Computer Vision and Pattern Recognition Conference (CVPR)},
  pages={12199--12208},
  year={2025},
  url={https://openaccess.thecvf.com/content/CVPR2025/papers/Wang_UniGraspTransformer_Simplified_Policy_Distillation_for_Scalable_Dexterous_Robotic_Grasping_CVPR_2025_paper.pdf}
}

@inproceedings{zhong2025dexgrasp,
  title={\href{https://openaccess.thecvf.com/content/CVPR2025/papers/Zhong_DexGrasp_Anything_Towards_Universal_Robotic_Dexterous_Grasping_with_Physics_Awareness_CVPR_2025_paper.pdf}{Dexgrasp anything: Towards universal robotic dexterous grasping with physics awareness}},
  author={Zhong, Yiming and Jiang, Qi and Yu, Jingyi and Ma, Yuexin},
  booktitle={Proceedings of the Computer Vision and Pattern Recognition Conference (CVPR)},
  pages={22584--22594},
  year={2025},
  url={https://openaccess.thecvf.com/content/CVPR2025/papers/Zhong_DexGrasp_Anything_Towards_Universal_Robotic_Dexterous_Grasping_with_Physics_Awareness_CVPR_2025_paper.pdf}
}

@ARTICLE{pavlichenko2025dexterous,
  author={Pavlichenko, Dmytro and Behnke, Sven},
  journal={IEEE Transactions on Automation Science and Engineering (TASE)},
  title={\href{https://doi.org/10.1109/TASE.2025.3541768}{Dexterous Pre-Grasp Manipulation for Human-Like Functional Categorical Grasping: Deep Reinforcement Learning and Grasp Representations}},
  year={2026},
  volume={23},
  number={},
  pages={2231--2244}, 
  doi={10.1109/TASE.2025.3541768},
  url={https://doi.org/10.1109/TASE.2025.3541768}
}

@inproceedings{attarian2023geometry,
  title={\href{https://arxiv.org/pdf/2312.03864}{Geometry matching for multi-embodiment grasping}},
  author={Attarian, Maria and Asif, Muhammad Adil and Liu, Jingzhou and Hari, Ruthrash and Garg, Animesh and Gilitschenski, Igor and Tompson, Jonathan},
  booktitle={Conference on Robot Learning (CORL)},
  pages={1242--1256},
  year={2023},
  url={https://arxiv.org/pdf/2312.03864}
}

@INPROCEEDINGS{li2022gendexgrasp,
  author={Li, Puhao and Liu, Tengyu and Li, Yuyang and Geng, Yiran and Zhu, Yixin and Yang, Yaodong and Huang, Siyuan},
  booktitle={IEEE International Conference on Robotics and Automation (ICRA)},
  title={\href{https://doi.org/10.1109/ICRA48891.2023.10160667}{GenDexGrasp: Generalizable Dexterous Grasping}},
  year={2023},
  volume={},
  number={},
  pages={8068--8074}, 
  doi={10.1109/ICRA48891.2023.10160667},
  url={https://doi.org/10.1109/ICRA48891.2023.10160667}
}

@INPROCEEDINGS{xu2024manifoundation,
  author={Xu, Zhixuan and Gao, Chongkai and Liu, Zixuan and Yang, Gang and Tie, Chenrui and Zheng, Haozhuo and Zhou, Haoyu and Peng, Weikun and Wang, Debang and Hu, Tianrun and Chen, Tianyi and Yu, Zhouliang and Shao, Lin},
  booktitle={IEEE/RSJ International Conference on Intelligent Robots and Systems (IROS)},
  title={\href{https://doi.org/10.1109/IROS58592.2024.10801782}{ManiFoundation Model for General-Purpose Robotic Manipulation of Contact Synthesis with Arbitrary Objects and Robots}},
  year={2024},
  volume={},
  number={},
  pages={10905--10912}, 
  doi={10.1109/IROS58592.2024.10801782},
  url={https://doi.org/10.1109/IROS58592.2024.10801782}
}

@inproceedings{robustdexgrasp,
    title={\href{https://arxiv.org/pdf/2504.05287}{{RobustDexGrasp}: Robust Dexterous Grasping of General Objects}},
    author={Zhang, Hui and Wu, Zijian and Huang, Linyi and Christen, Sammy and Song, Jie},
    booktitle={Conference on Robot Learning (CoRL)},
    year={2025},
    url={https://arxiv.org/pdf/2504.05287}
  }

@article{wei2025omnidexgrasp,
  title={OmniDexGrasp: Generalizable Dexterous Grasping via Foundation Model and Force Feedback},
  author={Wei, Yi-Lin and Luo, Zhexi and Lin, Yuhao and Lin, Mu and Liang, Zhizhao and Chen, Shuoyu and Zheng, Wei-Shi},
  journal={arXiv preprint arXiv:2510.23119},
  year={2025}
}

@article{wei2025cyclemanip,
  title={CycleManip: Enabling Cyclic Task Manipulation via Effective Historical Perception and Understanding},
  author={Wei, Yi-Lin and Liao, Haoran and Lin, Yuhao and Wang, Pengyue and Liang, Zhizhao and Liu, Guiliang and Zheng, Wei-Shi},
  journal={arXiv preprint arXiv:2512.01022},
  year={2025}
}

@article{wei2024grasp,
  title={Grasp as you say: Language-guided dexterous grasp generation},
  author={Wei, Yi-Lin and Jiang, Jian-Jian and Xing, Chengyi and Tan, Xian-Tuo and Wu, Xiao-Ming and Li, Hao and Cutkosky, Mark and Zheng, Wei-Shi},
  journal={Advances in Neural Information Processing Systems},
  volume={37},
  pages={46881--46907},
  year={2024}
}

@inproceedings{wei2025afforddexgrasp,
  title={Afforddexgrasp: Open-set language-guided dexterous grasp with generalizable-instructive affordance},
  author={Wei, Yi-Lin and Lin, Mu and Lin, Yuhao and Jiang, Jian-Jian and Wu, Xiao-Ming and Zeng, Ling-An and Zheng, Wei-Shi},
  booktitle={Proceedings of the IEEE/CVF International Conference on Computer Vision},
  pages={11818--11828},
  year={2025}
}

@article{lin2025typetele,
  title={Typetele: Releasing dexterity in teleoperation by dexterous manipulation types},
  author={Lin, Yuhao and Wei, Yi-Lin and Liao, Haoran and Lin, Mu and Xing, Chengyi and Li, Hao and Zhang, Dandan and Cutkosky, Mark and Zheng, Wei-Shi},
  journal={arXiv preprint arXiv:2507.01857},
  year={2025}
}

% ============================================================================
% Supplementary Material (template)
% This file is included at the end of paper_template.tex.
% We use S-numbering (S1, S2, ...) and fix hyperref anchors to avoid duplicates.
% ============================================================================

\clearpage

% ---- S-numbering for sections/subsections ----
\setcounter{section}{0}
\setcounter{subsection}{0}
\setcounter{subsubsection}{0}
\renewcommand{\thesection}{S\arabic{section}}
\renewcommand{\thesubsection}{S\arabic{section}.\arabic{subsection}}
\renewcommand{\thesubsubsection}{S\arabic{section}.\arabic{subsection}.\arabic{subsubsection}}

% ---- Fix hyperref anchor duplication when counters are reset ----
% IEEEtran + hyperref uses \theH<section> for PDF destinations.
% If we reset counters, we must also reset \theH<section> to a unique namespace.
\makeatletter
\providecommand{\theHsection}{}
\providecommand{\theHsubsection}{}
\providecommand{\theHsubsubsection}{}
\renewcommand{\theHsection}{supp.\arabic{section}}
\renewcommand{\theHsubsection}{supp.\arabic{section}.\arabic{subsection}}
\renewcommand{\theHsubsubsection}{supp.\arabic{section}.\arabic{subsection}.\arabic{subsubsection}}
\makeatother

\section*{\textbf{DexGrasp-Zero: A Morphology-Aligned Policy for Zero-Shot Cross-Embodiment Dexterous Grasping}}
\label{sec:supp_main}
\begin{center}
    {\large\bfseries Supplementary Material\par}
\end{center}

\noindent In Sec.~\ref{sec:supp_method_detail}, we provide additional method details, including morphology-aligned graph construction and the hand-specific primitive mapping $\mathcal{M}_h$.
In Sec.~\ref{sec:supp_impl_details}, we describe implementation details for MAGCN, PPO training, and sim-to-real deployment.
In Sec.~\ref{sec:supp_additional}, we report additional experiments and analyses, including training curves, per-object real-world results, representative failure cases, URDF-prior sensitivity, backbone comparisons, and zero-shot generalization to a non-anthropomorphic end-effector (Barrett Hand). \noindent \textbf{Most notably,} the policy trained on anthropomorphic hands transfers \emph{zero-shot} to the 8-DoF Barrett Hand and achieves a success rate of \textbf{0.70} on YCB objects in simulation, highlighting the extensibility of our morphology-aligned representation beyond anthropomorphic hands.

\section{Method Details}
\label{sec:supp_method_detail}

\subsection{Morphology-Aligned Graph Construction}
\label{subsec:supp_graph_features}
\begin{figure}[h]
    \centering
    \includegraphics[width=1\linewidth,trim=0cm 5.5cm 14cm 0cm,clip]{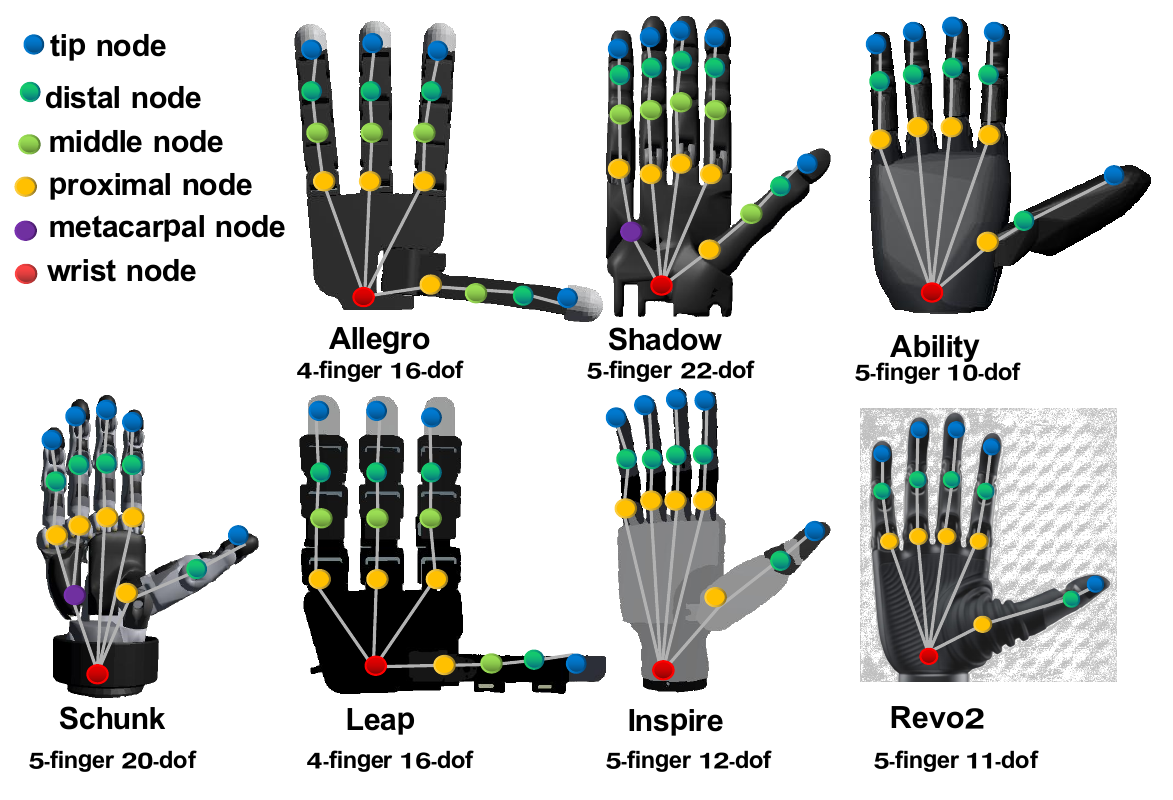}
    \caption{Morphology-aligned graph construction results for all hand embodiments used in this work.}
    \label{fig:supp_graph_all_hands}
\end{figure}

\begin{figure}[h]
    \centering
    \includegraphics[width=1\linewidth,trim=0cm 12cm 13cm 0cm,clip]{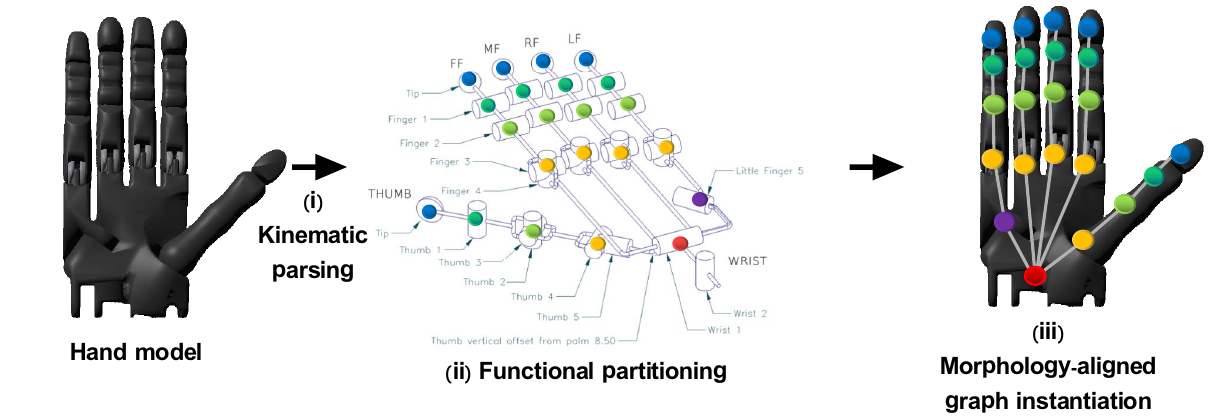}
    \caption{Example pipeline for morphology-aligned graph construction.}
    \label{fig:case}
\end{figure}
We represent each hand as a semantic graph whose nodes correspond to anatomical units (wrist, metacarpal, proximal, middle, distal, fingertip), and whose edges follow the kinematic tree.

Our construction consists of three stages.
(i) \textbf{URDF kinematic parsing}: we parse the URDF to obtain the joint--link kinematic tree, and extract the ordered joint chains for each finger branching from the wrist.
(ii) \textbf{Functional partitioning}: we assign each URDF joint to an anatomically meaningful semantic node (e.g., metacarpal/proximal/middle/distal/fingertip) based on its functional role along the finger chain.
A semantic node can contain one joint or multiple closely related joints.
In our implementation, each semantic node contains at most three joints, and each joint belongs to exactly one semantic node.
For example, on Allegro, \texttt{thumb4} and \texttt{thumb5} are assigned to the same \emph{thumb-proximal} node, as illustrated in Fig.~\ref{fig:case}.
(iii) \textbf{Morphology-aligned graph instantiation}: We treat semantic nodes above as the graph nodes and connect two semantic nodes with an edge if their underlying joint groups are adjacent along the URDF kinematic tree.

\paragraph{Special cases}
Some embodiments contain multiple joints that are spatially separated due to mechanical packaging, but correspond to the same functional region.
For example, on Allegro (see Fig.~\ref{fig:primitive_construct}), the two joints at the finger root have a noticeable physical offset, yet both contribute to the proximal-level articulation.
We therefore assign them to the same proximal semantic node and use the position of the first joint in the group as the node anchor.

\paragraph{Extensibility}
The semantic node set is not restricted to anthropomorphic hands.
For non-anthropomorphic end-effectors or special hand designs, as long as one or a group of URDF joints can be interpreted as a functional unit (e.g., a coupled finger, a parallel jaw, or a compliant module), we can define it as a semantic node and construct the corresponding morphology-aligned graph using the same pipeline.

\subsection{Mapping \texorpdfstring{$\mathcal{M}_h$}{M\_h} Construction}
\label{subsec:supp_mapping}

We use a fixed hand-specific mapping $\mathcal{M}_h$ to convert motion primitives into physical joint commands.
We implement $\mathcal{M}_h$ as an indexing rule: for the $j$-th physical DoF,
\begin{equation}
\Delta q_j^h = s_j^h\cdot \alpha_{n_j,(p_j)}^h.
\end{equation}

\begin{figure}[h]
    \centering
        \includegraphics[width=1\linewidth,trim=0.5cm 10cm 20cm 0cm,clip]{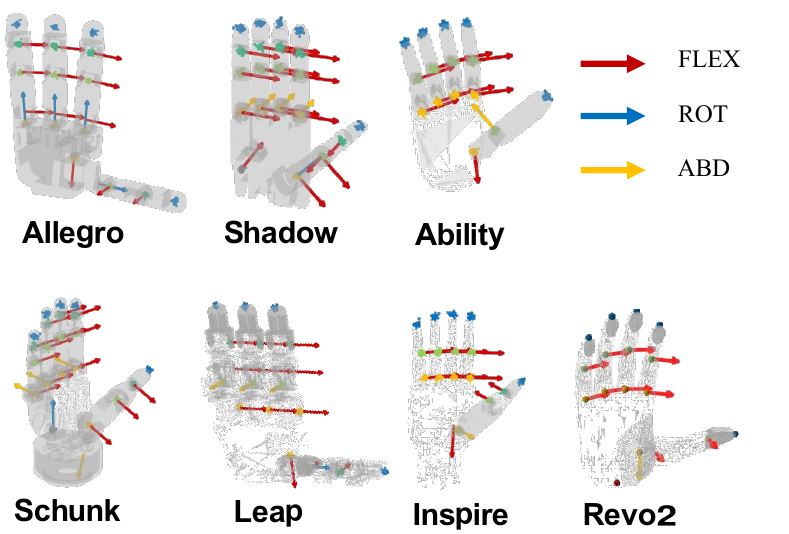}
    \caption{Joint-to-primitive mapping for all hand embodiments used in this work. Each URDF joint is assigned to one motion primitive.}
    \label{fig:primitive_construct}
\end{figure}

We construct $\mathcal{M}_h$ via a simulation-based system identification procedure.
For each actuated joint $j$, we apply a unit excitation to joint $j$ while keeping all other joints fixed, and measure the induced motion of the corresponding child unit in the palm frame.
The dominant response component determines the primitive type $p_j\in\{\text{FLEX},\text{ABD},\text{ROT}\}$, and the response direction determines the sign $s_j^h\in\{-1,+1\}$.
The node index $n_j$ denotes the semantic-node ID that joint $j$ is assigned to by the functional partitioning described in Sec.~\ref{subsec:supp_graph_features}.
This excitation--response analysis can be automated from URDF plus a physics engine, and it also supports manual annotation when preferred. 
Figure~\ref{fig:primitive_construct} visualizes the resulting joint-to-primitive assignments for all hands.

% \begin{figure}[h]
%     \centering
%         \includegraphics[width=1\linewidth,trim=0.cm 14cm 13.5cm 0cm,clip]{figs/fig-hands-compare2pdf.pdf}
%     \caption{Joint-to-primitive mapping for all hand embodiments used in this work. Each URDF joint is assigned to one motion primitive.}
%     \label{fig:primitive_construct}
% \end{figure}

\section{Implementation Details}
\label{sec:supp_impl_details}

\subsection{Benchmarks, Object Splits, and Real-World Props}
\label{subsec:supp_benchmarks_props}

\paragraph{Simulation benchmarks and object splits} %paragraph后面模板自带冒号，不需要写句号
In simulation, we strictly follow the official protocols and object splits from prior open-source benchmarks.
Specifically, we use (i) the CrossDex/YCB benchmark protocol~\cite{cross,YCB} and (ii) the GraspXL benchmark protocol~\cite{graspxl}.
We do not restate the full object lists here. Please refer to the original benchmark releases for the exact splits.

\begin{figure}[h]
    \centering
    \includegraphics[width=0.3\linewidth]{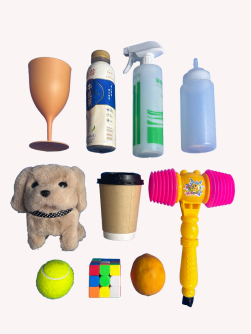}
    \caption{Real-world evaluation props (10 objects) used in our hardware experiments.}
    \label{fig:supp_obj_selected}
\end{figure}

\paragraph{Real-world props (10 objects)}
For real-world evaluation, we use a fixed set of 10 objects.
Figure~\ref{fig:supp_obj_selected} visualizes these objects.

The 10 objects are: (1) a wine-glass (plastic), (2) a beverage bottle, (3) a spray bottle, (4) a squeeze sauce bottle, (5) a toy dog, (6) a coffee mug, (7) a plastic toy hammer, (8) a tennis ball, (9) a Rubik's cube, and (10) an orange.
They span a wide range of sizes (from small objects such as the tennis ball with roughly 10~cm diameter, to larger objects such as the toy dog of about $20\!\times\!20\!\times\!20$~cm) and physical properties (from rigid objects to compliant/deformable ones).

\subsection{Observation Space Details}
\label{subsec:supp_obs_details}

Following RobustDexGrasp~\cite{robustdexgrasp}, we adopt a hand-centric observation representation.
Geometric terms, including the nearest-node-to-object distance vectors and the wrist-to-object vector, are expressed in the palm frame.
All observation channels are normalized to a comparable numeric scale for stable learning with fixed scaling and clipping for distance and force-related quantities.

\subsection{Policy Architecture (MAGCN)}
\label{subsec:supp_arch}

We use MAGCN with a 10-layer GCN encoder (hidden size 128 for the first layer and 256 for the remaining layers), ReLU activation and LayerNorm, and no dropout.
URDF priors are encoded by a lightweight MLP (one hidden layer with size 32 and LeakyReLU) and fused into the node stream, and a global MLP branch (hidden size 256, LeakyReLU) provides global context.
The actor outputs (i) a 6-DoF wrist command via an MLP head (hidden size 128, LeakyReLU) and (ii) three primitive scalars (FLEX/ABD/ROT) per node via another MLP head (hidden size 128, LeakyReLU), conditioned by a node-wise primitive-validity mask.
The critic shares the encoder and uses a value head (hidden size 128, ReLU) with scalar output.

\subsection{Training Details}
\label{subsec:supp_training_details}

Our RL implementation is based on GraspXL~\cite{graspxl}, and we train with PPO~\cite{ppo}.
Unless otherwise stated, we follow the training setup in the main paper. Table~\ref{tab:supp_hyperparams} lists the concrete hyperparameters used in our implementation.

\begin{table}[h]
    \centering
    \caption{Training and architecture hyperparameters.}
    \label{tab:supp_hyperparams}
    \setlength{\tabcolsep}{6pt}
    \renewcommand{\arraystretch}{1.1}
    \begin{tabular}{lc}
        \toprule
        \textbf{Parameter} & \textbf{Value} \\
        \midrule
        RL algorithm & PPO \\
        Discount factor $\gamma$ & 0.996 \\
        GAE parameter $\lambda$ & 0.95 \\
        Clipping parameter $\epsilon$ & 0.2 \\
        Optimizer & Adam \\
        Learning rate & $5\times 10^{-4}$ (adaptive KL schedule) \\
        Max gradient norm & 0.5 \\
        Batch size & $N_{\text{env}}\times 130$ transitions per update (4 mini-batches) \\
        PPO epochs / update & 4 \\
        Number of envs & $N_{\text{env}}=N_{\text{hand}}\times N_{\text{obj}}\times N_{\text{repeat}}$ (default $N_{\text{repeat}}{=}3$) \\
        Simulator frequency & 400~Hz physics / 20~Hz control \\
        Steps per episode & 120 + 30 lift steps \\ 
        Reward weights& $w_{\text{dis}}=0.3$, $w_{\text{contact}}=1.0$ \\
        &  $w_{\text{force}}=0.5$ ,$w_{\text{reg}}=1.5$ ,$w_{\text{pen}}=0.3$ \\
        \bottomrule
    \end{tabular}
\end{table}

At the beginning of each episode, we initialize the hand in an open-palm pose.
The wrist is randomly positioned at a distance of 30~cm from the object, with the palm oriented to face the object.
The object is initialized at the world origin with a random orientation.
Policy actions are executed by a PD controller with hand-agnostic gains: 0.015 for finger joints and 0.01 for arm joints. %Following GraspXL~\cite{graspxl}, we further incorporate an objective-driven guidance term with a gain of 0.05 that steers the hand toward the target pose, while the policy outputs only the residual action.

\subsection{Sim-to-Real}
\label{subsec:supp_sim2real}

We follow the teacher--student privileged distillation strategy from RobustDexGrasp~\cite{robustdexgrasp}.
We document the distillation details and the hardware deployment details separately.

\subsubsection{Privileged Distillation Details}
\label{subsubsec:supp_distill_details}

\paragraph{Teacher observations}
The privileged teacher policy has access to full visual-tactile information in simulation, including per-finger-link binary contact states and contact impulse magnitudes. These signals are not available on real hardware and are thus considered privileged.

\paragraph{Student architecture and history length}
The student policy shares the same MAGCN model as the teacher but replaces tactile inputs with temporal estimates. It uses a single-layer LSTM with hidden size 256 to process a history window of the last 5 time steps of proprioceptive and visual observations, enabling implicit contact reconstruction from motion residuals. 

\paragraph{Distillation objective}
We initialize the student with the teacher’s weights and train it via behavior cloning using mean squared error (MSE) between the student’s and teacher’s action outputs. No value function or policy entropy is distilled. Only the action imitation loss is used during the initial imitation phase before transitioning to reinforcement learning.

\subsubsection{Real-World Deployment Details}
\label{subsubsec:supp_hw_details}
\begin{itemize}
    \item \textbf{Perception pipeline.} We first capture an RGB-D image and use SAM2 to segment the object region in the RGB image and extract the corresponding object point cloud. The grasp target point is set to the 3D centroid of the segmented object point cloud (transformed into the robot base frame). Consistent with prior real-world setting in RobustDexGrasp~\cite{robustdexgrasp}, we assume the object remains stationary during the grasp. To reduce occlusion-induced noise, we reuse the point cloud from the initial perception frame as a fixed reference for all subsequent policy inferences.
    \item \textbf{Initialization.} After perception, we move the dexterous hand to a standardized pre-grasp pose relative to the target point: the palm faces the object and the hand starts with a fixed lateral offset of 25~cm. We additionally preset the thumb to an open pre-grasp configuration by setting its thumb-opening DoF to the maximum opening angle for each hand (LEAP, Inspire, and Revo2).
    \item \textbf{Control and safety.} We run policy inference at 20~Hz and clip the action outputs to safe ranges, using $0.015$ for finger commands and $0.01$ for arm commands. For wrist control, we use a high-frequency velocity controller, and apply a Kalman filter to the control signals to reduce jitter. We execute the grasp controller for 130 steps and then apply a scripted lift command.
\end{itemize}

\section{Additional Experiments and Analysis}
\label{sec:supp_additional}

\subsection{Training Curves in Simulation}
\label{subsec:supp_training_curve}
\begin{figure}[h]
    \centering
    \includegraphics[width=0.95\linewidth]{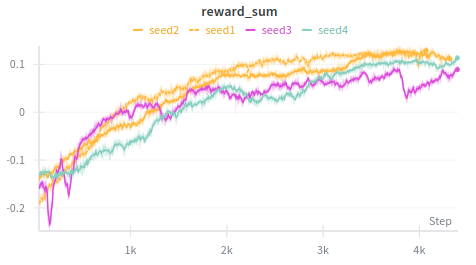}
    \caption{Training curves in simulation for our MAGCN policy.}
    \label{fig:supp_training_curve}
    \vspace{-1em}
\end{figure}

Figure~\ref{fig:supp_training_curve} shows that our training is stable across seeds and converges to consistent performance.
We plot the reward curves to reflect the policy improvement and the underlying PPO optimization dynamics.

\subsection{Per-Object Results in Real World}
\label{subsec:supp_per_object}
The object set is visualized in Fig.~\ref{fig:supp_obj_selected}. We report per-object success rates for each platform.

\begin{table}[h]
\centering
\caption{Per-object real-world results. Each object is evaluated over 5 poses per platform and we report successes out of 5; the average row reports the mean success rate.}
\label{tab:supp_per_object_real}
\setlength{\tabcolsep}{4pt}
\renewcommand{\arraystretch}{1.05}
\begin{tabular}{cccc}
\toprule
\textbf{Object} & \textbf{LEAP} & \textbf{Inspire} & \textbf{Revo2} \\
\midrule
wine-glass & 5/5 & 5/5 & 4/5 \\
Beverage bottle & 4/5 & 3/5 & 4/5 \\
Spray bottle & 4/5 & 4/5 & 4/5 \\
Squeeze sauce bottle & 4/5 & 5/5 & 4/5\\
Toy dog & 5/5 & 5/5 & 2/5 \\
Coffee mug & 5/5 & 5/5 & 4/5 \\
Plastic toy hammer & 4/5 & 5/5 & 2/5 \\
Tennis ball & 4/5 & 3/5 & 5/5 \\
Rubik's cube & 5/5 & 4/5 & 3/5 \\
Orange & 4/5 & 4/5 & 4/5 \\
\hline
Average & 0.88 & 0.86 & 0.72 \\
\bottomrule
\end{tabular}
\end{table}

\subsection{Failure Case Analysis}
\begin{figure}[h]
    \centering
    \IfFileExists{figs/fail_cases.pdf}{%
        \includegraphics[width=1\linewidth,trim=0.0cm 5.1cm 21.0cm 0cm,clip]{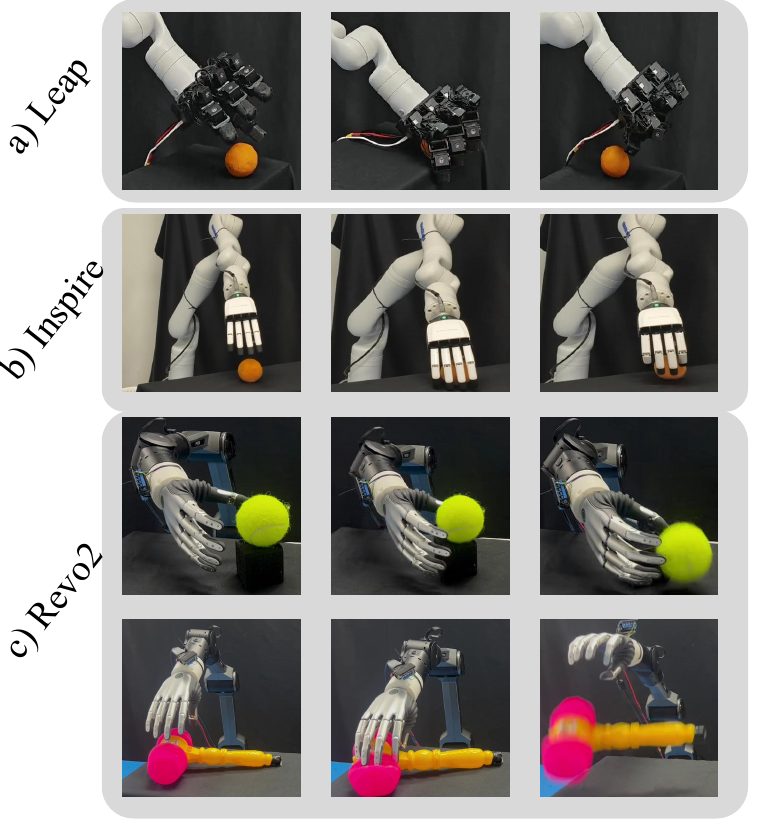}%
    }{%
        \fbox{\parbox{0.95\linewidth}{\centering Missing figure file: \texttt{figs/fail\_cases.pdf}}}%
    }
    \caption{Representative failure cases on the three unseen hands. (a) LEAP, (b) Inspire, (c) Revo2.}
    \label{fig:supp_fail_cases}
\end{figure}
We summarize representative failure modes observed in our real-world evaluation (Fig.~\ref{fig:supp_fail_cases}).
Overall, the dominant failures come from small objects, where missing tactile feedback makes it difficult to detect partial contact and recover from an unstable enclosure.

\textbf{(i) Small-object ``empty grasps'' on LEAP and Inspire.}
In Fig.~\ref{fig:supp_fail_cases}(a--b), for small objects the hand may close while the object is only partially captured, yielding an unstable enclosure that cannot survive lift-and-hold.
Without reliable tactile feedback on hardware, the policy cannot explicitly confirm secure contact formation and may lift prematurely.

\textbf{(ii) Embodiment-specific difficulty on Revo2.}
In Fig.~\ref{fig:supp_fail_cases}(c), Revo2's lower DoF and limited thumb dexterity can break thumb--finger opposition (tennis ball), while larger objects may exceed the hand's enclosure capacity (plastic toy hammer).

\subsection{URDF Prior Sensitivity}
\label{subsec:supp_urdf_sensitivity}
To verify that URDF physical priors are practically useful (beyond being additional inputs), we perform a controlled sensitivity test. We perturb only the URDF-derived feature encoding while keeping the trained policy, the simulator geometry/dynamics, and the evaluation protocol fixed.
Specifically, we scale every link length in the Allegro URDF by a constant factor $s\in\{1, 2, 1/4\}$, recompute the URDF-derived node features, and evaluate.
This isolates how the policy uses the encoded kinematic scale.

\begin{table}[h]
\centering
\caption{URDF prior sensitivity on Allegro via link-length scaling. We report success rate and qualitative behavior.}
\label{tab:supp_urdf_sensitivity}
\begin{tabular}{cc}
\toprule
\textbf{Scale} & \textbf{SR}  \\
\midrule
1 & 0.92   \\
2 & 0.85  \\
1/4 & 0.87  \\
\bottomrule
\end{tabular}
\end{table}

Table~\ref{tab:supp_urdf_sensitivity} shows that the success rate drops when the URDF scale deviates from the nominal setting, suggesting that the encoded kinematic scale is functionally used by the policy.
Qualitatively, we observe a consistent shift in enclosure timing.
With a larger scale ($s{=}2$), the hand starts closing earlier at a larger wrist-to-object distance, while with a smaller scale ($s{=}1/4$), it tends to close more gradually over the approach phase.

\subsection{Backbone Comparison}
\label{subsec:supp_backbone_compare}
We study whether the policy performance depends on the choice of graph backbone.
All comparisons use the same training pipeline and protocol; we change only the backbone architecture.
% We report success rate (SR, \%) and the standard deviation over three random seeds on \emph{train hands} and \emph{unseen hands}.

\textbf{Graph-Transformer hyperparameters.}
To keep the comparison fair, we match the overall model capacity to our GCN by using a comparable hidden width.
Unless otherwise noted, we use a 10-layer Graph-Transformer with $d_{model}{=}256$, 8 attention heads, dropout 0.1, and pre-layernorm (pre-LN).
The feed-forward dimension is 256 with GELU activations, and attention is applied on the same kinematic-tree-induced graph connectivity.
We also swept common design choices (e.g., dropout rate, feed-forward width, and minor architectural variants) and observed the same qualitative trend.

\begin{table}[h]
\centering
\caption{Backbone comparison (3 seeds). SR is reported on train hands and unseen hands.}
\label{tab:supp_backbone_compare}
\setlength{\tabcolsep}{4pt}
\renewcommand{\arraystretch}{1.12}
\begin{tabular}{lcc}
\toprule
\textbf{Backbone} & \textbf{Train-hand SR (\%)} & \textbf{Unseen-hand SR (\%)} \\
\midrule
GCN (ours) & 91.3/92.1/92.5 (91.9$\pm$0.6) & 86.1/84.9/85.5 (85.5$\pm$0.6) \\
Graph-Transformer & 86.4/87.2/86.9 (86.8$\pm$0.4) & 81.5/80.4/79.8 (80.6$\pm$0.9) \\
\bottomrule
\end{tabular}
\end{table}

As shown in Table~\ref{tab:supp_backbone_compare}, both graph backbones can solve the task and achieve high success rates.
However, GCN performs better and is more consistent across seeds on both train and unseen hands.
We attribute this to the stronger kinematic inductive bias and the simpler, local message-passing structure of GCN, which tends to be easier to optimize under RL and less sensitive to seed-level variation than attention-based backbones.

\subsection{Generalization to Non-Anthropomorphic End-Effectors}
\label{subsec:supp_non_anthro}
We further validate the universality of our representation on a non-anthropomorphic end-effector: the Barrett Hand, an 8-DoF three-finger gripper.
This experiment tests whether our morphology-aligned graph and motion-primitive interface can generalize beyond human-like kinematic layouts.

\begin{figure}[h]
    \centering
    \includegraphics[width=1\linewidth,trim=0.1cm 5.5cm 17.0cm 0cm,clip]{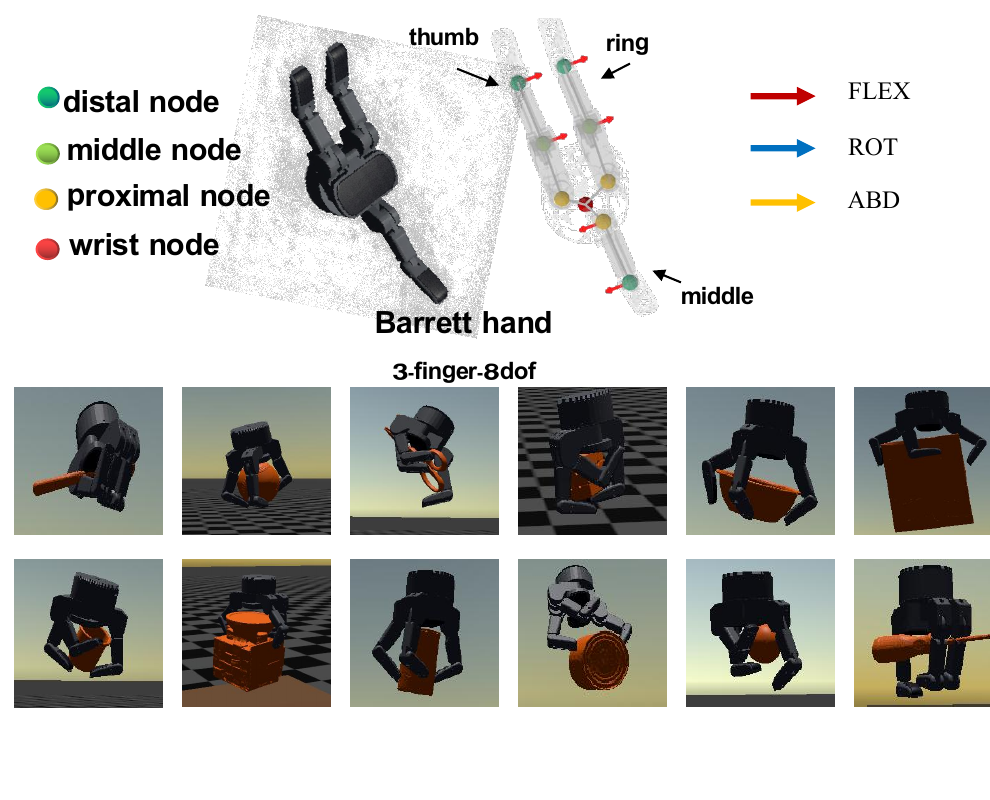}
    \caption{Barrett Hand graph construction and qualitative grasping examples.}
    \label{fig:barrett}
\end{figure}

\paragraph{Graph and mapping construction}
As illustrated in Fig.~\ref{fig:barrett}, we construct the hand graph by parsing the Barrett URDF and following its three-finger kinematic structure.
We define the three fingers as \emph{thumb}, \emph{middle}, and \emph{ring}, and instantiate semantic nodes according to the URDF-defined joint ordering and naming.
We then construct the hand-specific mapping $\mathcal{M}_h$ using the same unit-excitation procedure as in Sec.~\ref{subsec:supp_mapping}: each joint is individually excited in simulation to identify its dominant primitive type and sign.

\paragraph{Zero-shot deployment}
We directly deploy the policy trained on the four anthropomorphic hands in the main paper (\texttt{full\_model}) to the Barrett Hand in simulation, without any fine-tuning.
On the YCB benchmark objects, the zero-shot success rate is \textbf{0.70}.
For all runs, we initialize the gripper 0.30~m above the object with all joint angles set to zero (maximum opening for Barrett) and orient the gripper toward the object before the approach.

Overall, this result supports the feasibility of extending our framework to non-anthropomorphic grippers by changing only the URDF-derived graph and mapping, while reusing the same policy architecture and training recipe.

\end{document}